\documentclass{ifm-tr}

\usepackage[utf8]{inputenc}
\usepackage{amsmath,amsfonts,amssymb,mathtools}
\usepackage{colortbl}
\usepackage{enumitem}
\usepackage{fontawesome5}
\usepackage{tabularx}

\definecolor{loopfindingback}{HTML}{EAF3ED}
\definecolor{loopfindingtext}{HTML}{2F6B4F}
\definecolor{questionresultback}{HTML}{F0EEFF}
\definecolor{findinganalysisback}{HTML}{FFF3D6}
\definecolor{deltagaintext}{HTML}{008A45}
\definecolor{deltalosstext}{HTML}{D32F2F}

\newcolumntype{Y}{>{\raggedright\arraybackslash}X}

\usetikzlibrary{arrows.meta,calc,positioning}

\tcbset{
  paper summary box/.style={
    enhanced,
    breakable,
    frame hidden,
    sharp corners,
    boxrule=0pt,
    left=0.30cm,
    right=0.30cm,
    top=0.24cm,
    bottom=0.24cm,
    before skip=0.18cm,
    after skip=0.30cm,
  },
}

\newtcolorbox{questionresult}[1]{
  paper summary box,
  colback=questionresultback,
  before upper={\textbf{#1.}\enspace},
}

\newtcolorbox{findinganalysis}[1]{
  paper summary box,
  colback=findinganalysisback,
  before upper={\textbf{#1.}\enspace},
}

\newcommand{\findingstatement}[1]{%
  #1%
}

\newcommand{\analysissummary}[1]{%
  \par #1%
}

\graphicspath{{figs/}}
\hypersetup{pdftitle={Towards Looped Models Done Right: Part I - Topology, Input Injection, Recurrent-State Design},
  pdfauthor={Benhao Huang, Chufan Shi, Junlin Chen, Shicheng Wen, Zhengzhong Liu, Eric Xing, Xuezhe Ma}}

\title{Towards Looped Models Done Right}
\titlesubtitle{Part I: Topology, Input Injection, Recurrent-State Design}

\author[\(\ddagger\ast\)]{\href{https://huskydoge.github.io/}{Benhao Huang}}\authorsep{, }
\author[\(\ddagger\dagger\)]{Chufan Shi}\authorsep{, }
\author[\(\ddagger\)]{Junlin Chen}\authorsep{, }
\author[\(\ddagger\dagger\)]{Shicheng Wen}\authorsep{\\}
\author[\(\ddagger\)]{Zhengzhong Liu}\authorsep{, }
\author[\(\ddagger\)]{Eric Xing}\authorsep{, }
\author[\(\ddagger\dagger\)]{Xuezhe Ma}\authorsep{}

\affiliation[\(\ddagger\)]{Institute of Foundation Models}\affiliationsep{, }
\affiliation[\(\dagger\)]{USC}\affiliationsep{, }
\affiliation[\(\ast\)]{CMU}

\metadata[\faGithub\ Code]{\url{https://github.com/ifm-ai/xllm-loop}}
\date{July 31, 2026}

\abstract{
Looped language models reuse physical Transformer blocks across logical depth,
but established architectures entangle recurrence placement, input injection,
and recurrent state organization. We isolate these axes in models trained from
scratch at matched parameter scale, logical depth, and token budgets, and then
test whether the resulting design principles transfer to resident- and
active-parameter-matched Mixture-of-Experts (MoE) models.
\begin{itemize}[leftmargin=1.35em,itemsep=0.15em,topsep=0.25em,parsep=0pt]
\item \textbf{Loop design.} The prelude--recurrent block--coda sandwich architecture consistently improves performance on context extraction and reasoning-intensive tasks. Input injection primarily enhances knowledge retrieval and context utilization, whereas random initialization and separate H/L recurrent states provide mixed or negative returns. Taken together, these results favor the \textit{Huginn}-style design: at the 730M-parameter and 336B-token scale, the complete \textit{Huginn} architecture outperforms \textit{Ouro} across all ten benchmarks.

\item \textbf{MoE transfer.} The advantages of \textit{Huginn} over \textit{Ouro} transfer to Mixture-of-Experts models. At the 8BA0.8B (32BA3B when unrolled) scale trained on 500B tokens, \textit{Huginn} outperforms \textit{Ouro} on nearly all benchmarks spanning knowledge, reasoning, and coding, while also exhibiting more balanced expert routing.

\item \textbf{Apples-to-apples comparison.} Under matched training and inference FLOPs, the \textit{Huginn}-style model uses 75\% fewer resident parameters than a 112-layer feedforward MoE baseline. Despite its smaller physical parameters, it outperforms the feedforward baseline on DROP and GSM8K, matches it on MATH500, and remains competitive on the other benchmarks.
\end{itemize}
}

\begin{document}
\maketitle

\begin{center}
  \vspace{-1.0em}
  \includegraphics[width=0.98\textwidth]{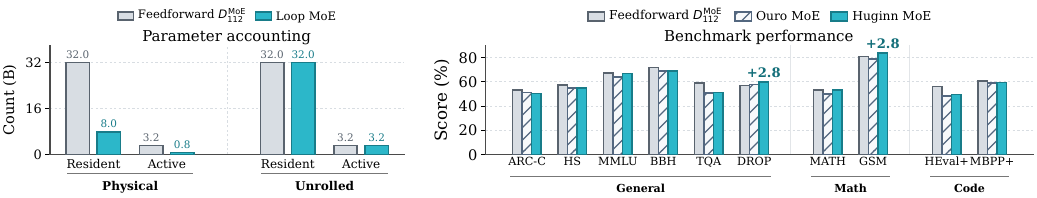}
  \captionsetup{type=figure,font=scriptsize,skip=2pt}
  \captionof{figure}{\textbf{Parameter scaling and benchmark performance of loop and feedforward MoE models.} \textbf{Left:} \textit{Ouro} MoE and \textit{Huginn} MoE share the same scale: 8.0B resident and 0.8B active parameters, corresponding to 32.0B resident-equivalent and 3.2B unrolled-active parameter applications. Unrolled counts measure parameter applications under weight reuse. \textbf{Right:} \textit{Huginn} MoE outperforms \textit{Ouro} MoE overall, and approaches or surpasses the 112-layer feedforward MoE baseline on DROP, MATH500 and GSM8K, while remaining competitive on the other benchmarks. All models use the same data and matched training and inference FLOPs, while looped models require less memory. Exact scores appear in Table~\ref{tab:allmoe-final-budget}.}
  \label{fig:overview-allmoe-benchmarks}
  \vspace{-1.0em}
\end{center}

\begingroup
\renewcommand{\contentsname}{Table of Contents}
\setcounter{tocdepth}{3}
\newpage
\tableofcontents
\endgroup

\section{Introduction}
\label{sec:introduction}

Looped language models repeatedly reuse the same Transformer blocks across logical depth, trading additional computation for reduced memory usage. As modern AI hardware delivers rapidly increasing compute throughput while memory capacity and bandwidth improve much more slowly~\citep{gholami2024aimemorywall}, this paradigm has emerged as a promising direction for scaling language models. Recent work has investigated recurrent-depth scaling~\citep{zhu2025ouro,geiping2025recurrentdepth,schwethelm2026recurrence,prairie2026parcae}, capability profiles~\citep{kohli2026loop,huang2026equilibrium}, as well as the roles of stability and residual dynamics~\citep{labovich2026stability,li2026deeploop}, architecture topology and sparsity~\citep{gao2026loopies,csordas2024moeut}, and memory efficiency~\citep{vendrell2026melt}. Most existing looped language models can be broadly categorized into two architectural lineages, represented by \textit{Ouro}~\citep{zhu2025ouro} and \textit{Huginn}~\citep{geiping2025recurrentdepth}.

Despite their growing influence, the architectural differences between these two families have not been systematically isolated. \textit{Ouro} recurrently applies a tied full stack initialized from token embeddings, whereas \textit{Huginn} places recurrence between untied prelude and coda layers, repeatedly injects a fixed prelude representation, and uses a separately initialized recurrent state. We formalize these differences along three axes---the iteration envelope, input injection, and latent-state organization---and isolate their effects through a controlled \textit{Ouro}-to-\textit{Huginn} transformation. We pretrain all resulting models from scratch under matched parameter scale, logical depth, and token budgets, and evaluate them using a common protocol.

\begin{figure}[!htbp]
  \centering
  \includegraphics[width=0.96\linewidth]{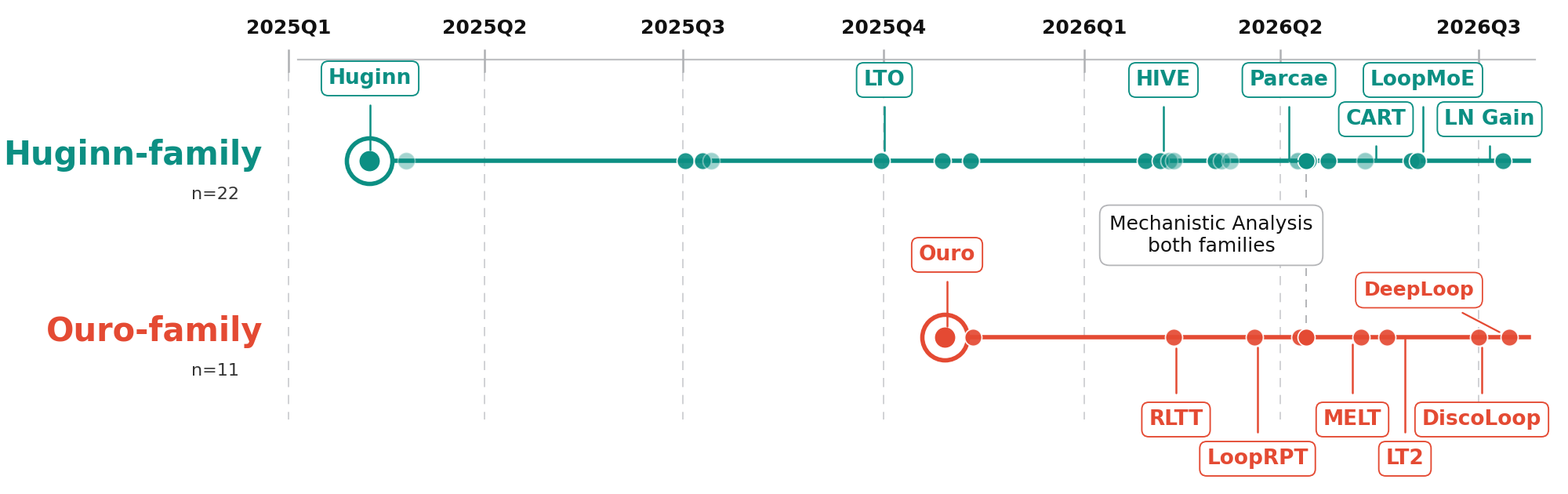}
  \caption{Selected looped-model works organized by family and quarter.
  Huginn-family:~\citep{geiping2025recurrentdepth,saunshi2025latentthoughts,lu2025latentcot,li2025skiploop,bae2025mixtureofrecursions,du2025lto,wu2025parallel,mcleish2025retrofitted,knupp2026dram,zhang2026hive,moosa2026dynamic,jeddi2026loopformer,song2026adaponderlm,chen2026moue,frey2026adaptiveloops,kohli2026loop,prairie2026parcae,zeitoun2026hyperloop,park2026loopus,capps2026cart,chen2026loopmoe,buehlmaier2026layernorm}.
  Ouro-family:~\citep{zhu2025ouro,fu2025thinkathard,williams2026rltt,tang2026looprpt,kirin2026relational,vendrell2026melt,deng2026lt2,fu2026discoloop,yu2026loopedssm,li2026deeploop,chen2026latentmt}.
  Cross-family mechanistic comparisons remain
  limited~\citep{blayney2026mechanistic}.
  Counts include all cited family-assigned papers\protect\footnotemark; the axis markers show a
  visual subset.}
  \label{fig:loop-family-timeline}
\end{figure}
\footnotetext{Literature cutoff: July 24, 2026.}

The rest of this article follows three questions.
\begin{enumerate}[leftmargin=1.8em]
\item \textbf{\hyperref[sec:preliminaries]{First, how should loop architectures be described?}}
We introduce a common framework that separates three choices often bundled
together: where recurrence is placed, how the input is reintroduced, and how
recurrent states are organized.

\item \textbf{\hyperref[sec:controlled-ablations]{Second, which of these choices actually matter?}}
Through controlled \textit{Ouro}-to-\textit{Huginn} ablations, we find that
placing recurrence inside a prelude--loop--coda sandwich provides the clearest
gains, especially on mathematical reasoning, context extraction, and other
reasoning-intensive tasks. Input injection helps models recover
prompt-provided context and specifications, but can interfere with
quantitative reasoning. Random initialization offers no consistent advantage,
and separate high/low states add complexity without reliable gains. When these
choices are combined, the full \textit{Huginn} design outperforms
\textit{Ouro} across all ten dense benchmarks.

\item \textbf{\hyperref[sec:moe-transfer]{Third, do these conclusions survive in MoE models?}}
They do. \textit{Huginn} again outperforms \textit{Ouro}, distributes expert
load more evenly, and uses different experts meaningfully across loop
iterations. It also approaches the performance of a much larger feedforward
MoE while storing substantially fewer parameters.
\end{enumerate}

\section{Preliminary}
\label{sec:preliminaries}

The architecture of \textit{Ouro} and \textit{Huginn} differs in their
iteration envelope, input injection, and recurrent-state organization.

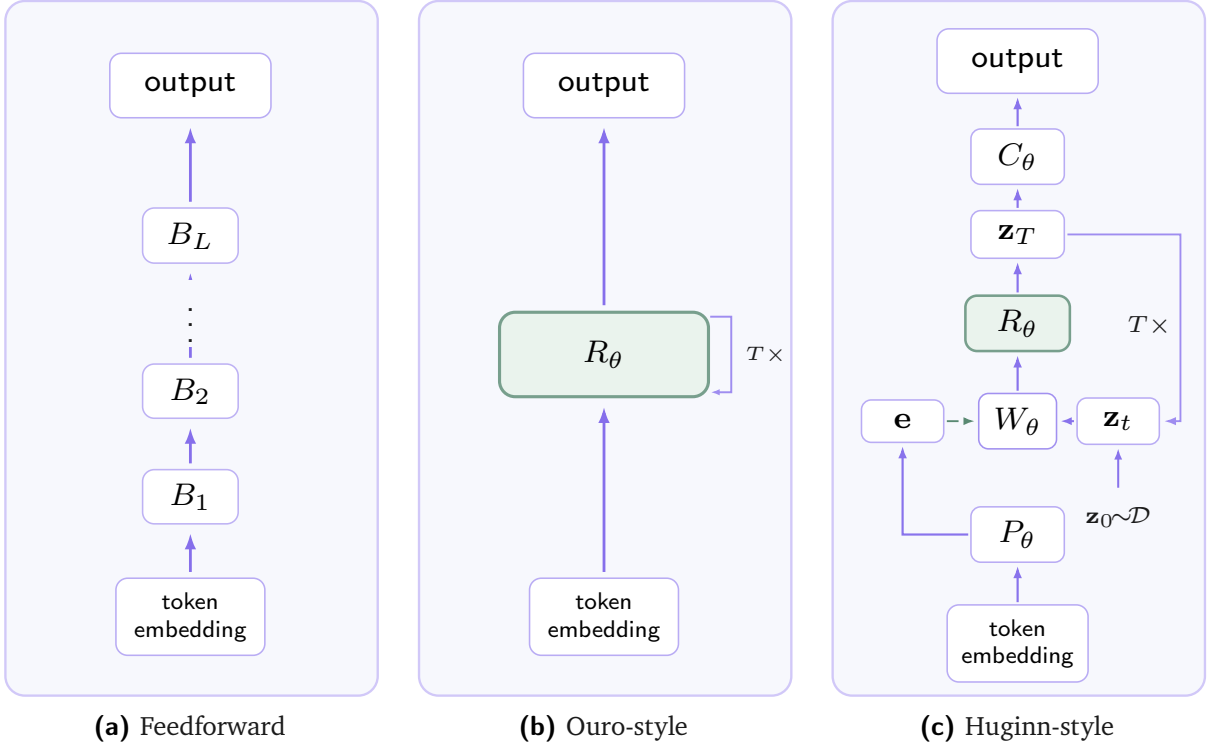
\begin{figure}[!htbp]
  \centering
  \captionsetup[subfigure]{justification=centering, font=small, labelfont=bf, labelsep=space}
  \tikzset{
    arch panel/.style={draw=ifmpurpletext!30, fill=ifmblueaccent!26, rounded corners=5pt, line width=0.6pt},
    arch box/.style={draw=ifmpurpletext!45, fill=white, rounded corners=3pt, minimum width=1.45cm, minimum height=0.58cm, align=center},
    arch token/.style={draw=ifmpurpletext!45, fill=white, rounded corners=3pt, minimum width=1.34cm, minimum height=0.52cm, align=center, font=\sffamily\tiny},
    arch small/.style={draw=ifmpurpletext!42, fill=white, rounded corners=3pt, minimum width=0.86cm, minimum height=0.44cm, align=center},
    arch recurrent/.style={draw=loopfindingtext!65, fill=loopfindingback, rounded corners=4pt, minimum width=1.88cm, minimum height=0.76cm, align=center, line width=0.8pt},
    arch hbox/.style={draw=ifmpurpletext!45, fill=white, rounded corners=2.5pt, minimum width=0.84cm, minimum height=0.34cm, align=center, font=\scriptsize},
    arch htied/.style={draw=loopfindingtext!65, fill=loopfindingback, rounded corners=3pt, minimum width=0.94cm, minimum height=0.38cm, align=center, line width=0.66pt, font=\scriptsize},
    arch write/.style={draw=ifmpurpletext!60, fill=white, rounded corners=2.5pt, minimum width=0.70cm, minimum height=0.34cm, align=center, font=\scriptsize},
    arch flow/.style={-{Latex[length=1.45mm,width=1.05mm]}, line width=0.72pt, draw=ifmpurpletext!78, shorten >=2.2pt, shorten <=1.4pt},
    arch loop/.style={-{Latex[length=0.82mm,width=0.58mm]}, line width=0.50pt, draw=ifmpurpletext!62, shorten >=0.4pt, shorten <=0.3pt},
    arch hflow/.style={-{Latex[length=1.12mm,width=0.80mm]}, line width=0.56pt, draw=ifmpurpletext!76, shorten >=0.7pt, shorten <=0.5pt},
    arch hloop/.style={-{Latex[length=1.08mm,width=0.76mm]}, line width=0.48pt, draw=ifmpurpletext!62, shorten >=1.0pt, shorten <=0.6pt},
    arch hshort/.style={-{Latex[length=1.02mm,width=0.74mm]}, line width=0.52pt, draw=ifmpurpletext!76, shorten >=0.8pt, shorten <=0.6pt},
    arch hshort inject/.style={-{Latex[length=1.02mm,width=0.74mm]}, dashed, line width=0.48pt, draw=loopfindingtext!80, shorten >=0.8pt, shorten <=0.6pt},
    arch note/.style={font=\sffamily\scriptsize, text=ifmtext, align=center},
  }

  \begin{subfigure}[t]{0.28\linewidth}
    \centering
    \resizebox{\linewidth}{!}{%
    \begin{tikzpicture}[font=\sffamily\scriptsize, >=Latex]
      \path[use as bounding box] (0,0) rectangle (3.35,6.25);
      \node[arch panel, minimum width=3.35cm, minimum height=6.25cm, anchor=south west] at (0,0) {};
      \node[arch token] (femb) at (1.675,0.72) {token\\embedding};
      \node[arch small] (f1) at (1.675,1.80) {$B_1$};
      \node[arch small] (f2) at (1.675,2.75) {$B_2$};
      \coordinate (fdots) at (1.675,3.45);
      \node[arch small] (fL) at (1.675,4.15) {$B_L$};
      \node[arch box] (fout) at (1.675,5.50) {output};
      \draw[arch flow] (femb.north) -- (f1.south);
      \draw[arch flow] (f1.north) -- (f2.south);
      \draw[arch flow] (f2.north) -- (fL.south);
      \node[arch note, fill=ifmblueaccent!26, inner sep=1.2pt] at (fdots) {$\vdots$};
      \draw[arch flow] (fL.north) -- (fout.south);
    \end{tikzpicture}%
    }
    \caption{Feedforward}
    \label{fig:architecture-comparison-feedforward}
  \end{subfigure}\hspace{0.025\linewidth}
  \begin{subfigure}[t]{0.28\linewidth}
    \centering
    \resizebox{\linewidth}{!}{%
    \begin{tikzpicture}[font=\sffamily\scriptsize, >=Latex]
      \path[use as bounding box] (0,0) rectangle (3.35,6.25);
      \node[arch panel, minimum width=3.35cm, minimum height=6.25cm, anchor=south west] at (0,0) {};
      \node[arch token] (oemb) at (1.675,0.72) {token\\embedding};
      \node[arch recurrent] (or) at (1.675,3.08) {$R_\theta$};
      \node[arch box] (out) at (1.675,5.50) {output};
      \draw[arch flow] (oemb.north) -- (or.south);
      \draw[arch flow] (or.north) -- (out.south);
      \draw[arch loop] (2.62,3.42) -- (2.82,3.42) -- (2.82,2.74) -- (2.62,2.74);
      \node[arch note, font=\tiny, anchor=west, inner sep=0.3pt] at (2.93,3.08) {$\scriptscriptstyle T{\times}$};
    \end{tikzpicture}%
    }
    \caption{Ouro-style}
    \label{fig:architecture-comparison-ouro}
  \end{subfigure}\hspace{0.025\linewidth}
  \begin{subfigure}[t]{0.28\linewidth}
    \centering
    \resizebox{\linewidth}{!}{%
    \begin{tikzpicture}[font=\sffamily\scriptsize, >=Latex]
      \path[use as bounding box] (0,0) rectangle (3.35,6.25);
      \node[arch panel, minimum width=3.35cm, minimum height=6.25cm, anchor=south west] at (0,0) {};
      \node[arch token, minimum width=1.04cm, minimum height=0.42cm] (hemb) at (1.675,0.48) {token\\embedding};
      \node[arch hbox] (pre) at (1.675,1.46) {$P_\theta$};
      \node[arch hbox, minimum width=0.74cm] (emb) at (0.64,2.48) {$\mathbf e$};
      \node[arch hbox, minimum width=0.74cm] (zin) at (2.58,2.48) {$\mathbf z_t$};
      \coordinate (zinit) at (2.58,1.86);
      \node[arch note, font=\scriptsize] at (2.58,1.62)
        {$\scriptstyle \mathbf z_0\!\sim\!\mathcal{D}$};
      \node[arch write] (hplus) at (1.675,2.48) {$W_\theta$};
      \node[arch htied] (core) at (1.675,3.36) {$R_\theta$};
      \node[arch hbox] (zout) at (1.675,4.16) {$\mathbf z_T$};
      \node[arch hbox] (coda) at (1.675,4.86) {$C_\theta$};
      \node[arch box] (hout) at (1.675,5.72) {output};
      \draw[arch hflow] (hemb.north) -- (pre.south);
      \draw[arch hflow] (pre.west) -| (emb.south);
      \draw[arch hshort] (zinit) -- (zin.south);
      \draw[arch hshort] (zin.west) -- (hplus.east);
      \draw[arch hflow] (hplus.north) -- (core.south);
      \draw[arch hflow] (core.north) -- (zout.south);
      \draw[arch hflow] (zout.north) -- (coda.south);
      \draw[arch hflow] (coda.north) -- (hout.south);
      \draw[arch hloop] (zout.east) -- (3.14,4.16) -- (3.14,2.48) -- (zin.east);
      \node[arch note, font=\scriptsize] at (2.86,3.34) {$\scriptstyle T{\times}$};
      \draw[arch hshort inject] (emb.east) -- (hplus.west);
    \end{tikzpicture}%
    }
    \caption{Huginn-style}
    \label{fig:architecture-comparison-huginn}
  \end{subfigure}

  \caption{Architecture comparisons. Feedforward models use distinct blocks.
  Ouro-style models reuse a full Transformer stack. Huginn-like models
  initialize the recurrent hidden stream from a prior, place a tied iterative
  body between untied prelude and coda stacks, and inject the prelude
  representation into the iterations.}
  \label{fig:architecture-comparison}
\end{figure}

\subsection{Notations}

Let $\mathbf x\in\mathcal V^{B\times L}$ denote token indices and
$\mathbf x_0=E_\theta(\mathbf x)\in\mathbb R^{B\times L\times d}$ their
embeddings. A general tied iterative model can be written as
\begin{align}
\mathbf e &= P_\theta(\mathbf x_0), \\
\mathbf z_0 &= \phi_\theta(\mathbf e,\boldsymbol{\xi}), \\
\tilde{\mathbf z}_t &= W_\theta(\mathbf z_t,\mathbf e), \\
\mathbf z_{t+1} &= R_\theta(\tilde{\mathbf z}_t), \qquad t=0,\ldots,T-1, \\
\mathbf h &= C_\theta(\mathbf z_T).
\end{align}
Here $P_\theta$ is the prelude map~\citep{geiping2025recurrentdepth},
$R_\theta$ is the tied iterative body, and $C_\theta$ is the coda map. All
three are stacks of standard Transformer blocks; the prelude and coda are
applied once, whereas the recurrent core is repeatedly applied with shared
parameters. $\phi_\theta$ initializes the recurrent state, and $W_\theta$ is
the per-step input write. The fixed representation $\mathbf e$ is the prelude
output, $\mathbf z_t$ is the model's recurrent hidden stream, and auxiliary
randomness $\boldsymbol{\xi}$ is used only by variants with random state
initialization.

\subsection{Comparing \textit{Ouro} and \textit{Huginn}}

Under this notation, an \textit{Ouro}-style topology sets $P_\theta$,
$W_\theta$, and $C_\theta$ to identity maps, initializes
$\mathbf z_0=\mathbf x_0$, and applies the same stack for $T$ iterations.

A \textit{Huginn}-style topology keeps nontrivial prelude and coda maps and
writes the fixed representation $\mathbf e$ into the recurrent stream at each
iteration. Note that such random initialization changes the starting value of
that same stream while leaving the number of persistent states unchanged.

\begin{table}[!htbp]
  \centering
  \footnotesize
  \setlength{\tabcolsep}{5pt}
  \renewcommand{\arraystretch}{1.18}
  \caption{\textit{Ouro}- and \textit{Huginn}-style instantiations of the
  common iterative-model notation, when given a layer budget of 112. Each row isolates one architectural
  choice.}
  \label{tab:ouro-huginn-symbol-map}
  \begin{tabularx}{0.96\linewidth}{
    @{}
    >{\raggedright\arraybackslash}p{0.22\linewidth}
    >{\raggedright\arraybackslash}X
    >{\raggedright\arraybackslash}X
    @{}}
    \toprule
    \rowcolor{ifmblueaccent}
    \textbf{Model specification}
      & \textbf{\textit{Ouro}-style}
      & \textbf{\textit{Huginn}-style} \\
    \midrule
    Prelude $P_\theta$
      & identity
      & 8 untied layers \\
    Initializer $\phi_\theta$
      & $\mathbf z_0=\mathbf x_0$
      & $\mathbf z_0=\boldsymbol{\xi}$ \\
    Input write $W_\theta$
      & $W_\theta(\mathbf z_t,\mathbf e)=\mathbf z_t$
      & $\boldsymbol\alpha\odot\mathbf z_t
         +\boldsymbol\delta\odot(\mathbf W_{\mathrm{in}}\mathbf e)$ \\
    Shared body $R_\theta$
      & tied 28-block stack
      & tied 12-block middle body \\
    Coda $C_\theta$
      & identity
      & 8 untied layers \\
    Iterations $T$
      & 4
      & 8 \\
    Layer Budgets     & 112 = 28 $\times$ 4
      & 112 = 8 + 12 $\times$ 8 + 8\\
    Resident Layers       & 28 
      & 28 = 8 + 12 + 8 \\
    \bottomrule
  \end{tabularx}
\end{table}

This decomposition yields three separable design axes:

\begin{itemize}[leftmargin=1.8em]
\item \textbf{Iteration envelope} chooses the placement and sharing pattern of
  $P_\theta,R_\theta,C_\theta$
\item \textbf{Input injection} chooses whether $W_\theta$ receives a persistent
  input condition and what it reads, e.g.\ token embeddings versus
  prelude-encoded $\mathbf e$;
\item \textbf{Latent-state organization} chooses how $\phi_\theta$ initializes
  the loop state, e.g.\ $\mathbf z_0=\mathbf e$ versus a separate draw,
  as well as the structures of the states, e.g.\ fast and slow latents.
\end{itemize}

\section{Controlled Ablations of Looped-Model Design}
\label{sec:controlled-ablations}

In this section, we specify the matched model, training, and evaluation controls, then report the stepwise ablations and the complete construction path from \textit{Ouro} to \textit{Huginn}. Q1 changes the iteration envelope, Q2 adds the prelude-conditioned input injection, and Q3 varies latent-state initialization and organization.

\subsection{Model Configuration}

We evaluate all looped-model variants at a matched physical scale of 730M
stored parameters and an equivalent unrolled scale of 2.9B parameter
applications. Each looped model stores the parameters of 28 Transformer blocks
but reuses a subset of these blocks across recurrent iterations, yielding an
unrolled computation path with the same logical depth as a 112-layer
feedforward Transformer.

\begin{itemize}[leftmargin=1.8em]
\item The \textit{Ouro}-style loop ties the full 28-block stack and executes it four times, giving $(28 \times 4 = 112)$ logical block executions, abbreviated as $R_{28}^4$.
\item The \textit{Huginn}-style loop instead places recurrence in the middle of a feedforward envelope: an 8-layer prelude, an 8-step loop over a 12-layer tied recurrent core, and an 8-layer coda, giving $(8 + 12 \times 8 + 8 = 112)$ logical block executions, abbreviated as $P_8 R_{12}^8 C_8$.
\end{itemize}

Two feedforward baselines anchor the comparison. The 730M baseline $D_{28}$
contains 28 Transformer blocks and matches the resident parameter count of the
looped models. The 2.9B baseline $D_{112}$ contains 112 independently
parameterized Transformer blocks and matches their unrolled logical depth.

\textbf{Transformer block.} All models use the same decoder-only PreNorm
backbone, ensuring that the ablations modify the recurrence pattern rather than
the local block architecture. The attention module uses 24 query heads, 6
key--value heads, a head dimension of 64, and RoPE with base $10^{6}$. At a
hidden dimension of 1536, the dense feed-forward branch uses a SwiGLU MLP with
intermediate width 4352. In each MoE variant, the attention, normalization,
and residual structure remain unchanged; only the dense feed-forward branch is
replaced by a top-2 routed mixture of experts.
Appendix~\ref{app:model-training} gives the complete configuration.

\subsection{Training Recipe}

All models are trained on the
\href{https://huggingface.co/spaces/LLM360/TxT360}{\textbf{TxT360 dataset}}~\citep{txt360data2024},
a high-quality dataset that deduplicates 99 Common Crawl snapshots and 14
curated data sources from diverse domains. We use sequence length 8192 and an
effective global batch of 512 sequences, or 4,194,304 tokens per optimizer
step. We optimize with AdamW, sweep learning rates in
$\{4\times10^{-4},6\times10^{-4},8\times10^{-4}\}$, and report the best
setting. We use 200 warmup steps followed by cosine decay and set weight decay
to 0.1.

\subsection{Evaluation Protocols}

We evaluate the models on ten benchmarks spanning several capability groups:
\begin{itemize}[leftmargin=1.8em]
\item ARC-Challenge, HellaSwag, MMLU, and TriviaQA emphasize factual,
  conceptual, and commonsense knowledge~\citep{clark2018arc,zellers2019hellaswag,hendrycks2021mmlu,joshi2017triviaqa};
\item BBH-CoT and DROP require reasoning over instructions, demonstrations, or
  passage-level evidence~\citep{suzgun2023challenging,dua2019drop};
\item GSM8K and MATH500 focus on multi-step quantitative and symbolic
  reasoning~\citep{cobbe2021training,hendrycks2021math,lightman2024verify}; and
\item HumanEval$^+$ and MBPP$^+$ test code generation under explicit functional
  specifications~\citep{chen2021evaluating,austin2021program,liu2023evalplus}.
\end{itemize}
These categories overlap, but they provide a useful lens for understanding
which capabilities benefit from each architectural change.
Appendix~\ref{app:evaluation-protocol} gives the splits, prompts, metrics, and
length caps.

\subsection{Controlled Architecture Ablations}

\begin{figure}[!htbp]
  \centering
  \resizebox{0.92\linewidth}{!}{%
  \begin{tikzpicture}[
    font=\sffamily\scriptsize,
    >=Latex,
    arch panel/.style={
      draw=ifmpurpletext!30,
      fill=ifmblueaccent!26,
      rounded corners=5pt,
      line width=0.6pt
    },
    arch box/.style={
      draw=ifmpurpletext!45,
      fill=white,
      rounded corners=3pt,
      minimum width=1.45cm,
      minimum height=0.58cm,
      align=center
    },
    arch token/.style={
      draw=ifmpurpletext!45,
      fill=white,
      rounded corners=3pt,
      minimum width=1.34cm,
      minimum height=0.52cm,
      align=center,
      font=\sffamily\tiny
    },
    arch small/.style={
      draw=ifmpurpletext!45,
      fill=white,
      rounded corners=2.5pt,
      minimum width=0.84cm,
      minimum height=0.34cm,
      align=center
    },
    arch recurrent/.style={
      draw=loopfindingtext!65,
      fill=loopfindingback,
      rounded corners=3pt,
      minimum width=1.20cm,
      minimum height=0.48cm,
      align=center,
      line width=0.7pt
    },
    arch write/.style={
      draw=ifmpurpletext!60,
      fill=white,
      rounded corners=2.5pt,
      minimum width=0.72cm,
      minimum height=0.36cm,
      align=center
    },
    arch flow/.style={
      -{Latex[length=1.25mm,width=0.92mm]},
      line width=0.72pt,
      draw=ifmpurpletext!78,
      shorten >=3.2pt,
      shorten <=2.2pt
    },
    arch loop/.style={
      -{Latex[length=0.82mm,width=0.58mm]},
      line width=0.50pt,
      draw=ifmpurpletext!62,
      shorten >=0.4pt,
      shorten <=0.3pt
    },
    arch inject/.style={
      -{Latex[length=1.00mm,width=0.72mm]},
      line width=0.56pt,
      draw=loopfindingtext!82,
      shorten >=2.4pt,
      shorten <=1.6pt
    },
    arch note/.style={
      font=\sffamily\scriptsize,
      text=ifmtext,
      align=center
    },
    arch title/.style={
      font=\sffamily\bfseries\tiny,
      text=ifmtext,
      align=center
    }
  ]
    \def\panelw{3.35}
    \def\panelh{6.30}

    \begin{scope}[shift={(0,0.72)}]
      \node[
        arch panel,
        minimum width=\panelw cm,
        minimum height=\panelh cm,
        anchor=south west
      ] at (0,0) {};
      \node[arch token] (femb) at (1.675,0.72) {token\\embedding};
      \node[
        arch recurrent,
        minimum width=1.95cm,
        minimum height=0.82cm
      ] (fullr) at (1.675,3.20) {$R_\theta$};
      \node[arch box] (fout) at (1.675,5.70) {output};
      \draw[arch flow] (femb.north) -- (fullr.south);
      \draw[arch flow] (fullr.north) -- (fout.south);
      \draw[arch loop, rounded corners=1.5pt]
        (fullr.north east) -- ++(0.28,0) |- (fullr.south east);
      \node[arch note, font=\tiny, anchor=west, inner sep=0.3pt]
        at (3.02,3.20) {$4{\times}$};
    \end{scope}

    \draw[arch flow, line width=0.9pt, shorten >=1pt, shorten <=1pt]
      (3.70,3.52) -- (4.48,3.52);

    \begin{scope}[shift={(4.83,0.72)}]
      \node[
        arch panel,
        minimum width=\panelw cm,
        minimum height=\panelh cm,
        anchor=south west
      ] at (0,0) {};
      \node[arch token] (memb) at (1.675,0.55) {token\\embedding};
      \node[arch small] (mpre) at (1.675,1.62) {$P_\theta$};
      \node[arch recurrent] (mcore) at (1.675,3.20) {$R_\theta$};
      \node[arch small] (mcoda) at (1.675,4.48) {$C_\theta$};
      \node[arch box] (mout) at (1.675,5.70) {output};
      \draw[arch flow] (memb.north) -- (mpre.south);
      \draw[arch flow] (mpre.north) -- (mcore.south);
      \draw[arch flow] (mcore.north) -- (mcoda.south);
      \draw[arch flow] (mcoda.north) -- (mout.south);
      \draw[arch loop]
        (2.43,3.54) -- (2.64,3.54) -- (2.64,2.86) -- (2.43,2.86);
      \node[arch note, font=\tiny, anchor=west, inner sep=0.3pt]
        at (2.75,3.20) {$8{\times}$};
    \end{scope}

    \draw[arch flow, line width=0.9pt, shorten >=1pt, shorten <=1pt]
      (8.53,3.52) -- (9.31,3.52);

    \begin{scope}[shift={(9.66,0.72)}]
      \node[
        arch panel,
        minimum width=\panelw cm,
        minimum height=\panelh cm,
        anchor=south west
      ] at (0,0) {};
      \node[arch token] (iemb) at (1.675,0.40) {token\\embedding};
      \node[arch small] (ipre) at (1.675,1.36) {$P_\theta$};
      \node[arch small, minimum width=0.62cm] (estate) at (0.68,1.98)
        {$\mathbf e$};
      \node[arch small, minimum width=0.66cm] (hstate) at (2.58,1.98)
        {$\mathbf z_t$};
      \node[arch note, font=\tiny] at (2.58,1.54)
        {$\mathbf z_0=\mathbf e$};
      \node[arch write] (write) at (1.675,2.78) {$W_\theta$};
      \node[arch recurrent] (icore) at (1.675,3.78) {$R_\theta$};
      \node[arch small] (icoda) at (1.675,4.72) {$C_\theta$};
      \node[arch box] (iout) at (1.675,5.70) {output};
      \draw[arch flow] (iemb.north) -- (ipre.south);
      \draw[arch flow, shorten >=1.2pt, shorten <=1.0pt]
        (ipre.west) -- (0.68,1.36) -- (estate.south);
      \draw[arch inject] (estate.north east) -- (write.south west);
      \draw[arch inject, draw=ifmpurpletext!78]
        (hstate.north west) -- (write.south east);
      \draw[arch flow] (write.north) -- (icore.south);
      \draw[arch flow] (icore.north) -- (icoda.south);
      \draw[arch flow] (icoda.north) -- (iout.south);
      \draw[arch loop]
        (icore.east) -- (3.17,3.78) -- (3.17,1.98) -- (hstate.east);
      \node[arch note, font=\tiny, anchor=west, inner sep=0.3pt]
        at (2.60,4.16) {$8{\times}$};
    \end{scope}

    \node[arch title] at (1.675,0.25) {(a) Full-stack loop};
    \node[arch title] at (6.505,0.25) {(b) Middle loop (sandwich)};
    \node[arch title] at (11.335,0.25)
      {(c) Middle loop + input injection};
  \end{tikzpicture}%
  }
  \caption{The controlled architecture ablations in Q1 \& Q2.
  Q1 moves recurrence from the full stack into a sandwich loop between untied
  prelude and coda layers; Q2 then further injects
  the evolving $\mathbf z_t$
  with the fixed prelude embedding $\mathbf e$ before every recurrent-core
  iteration.}
  \label{fig:q1-q2-architecture-path}
\end{figure}
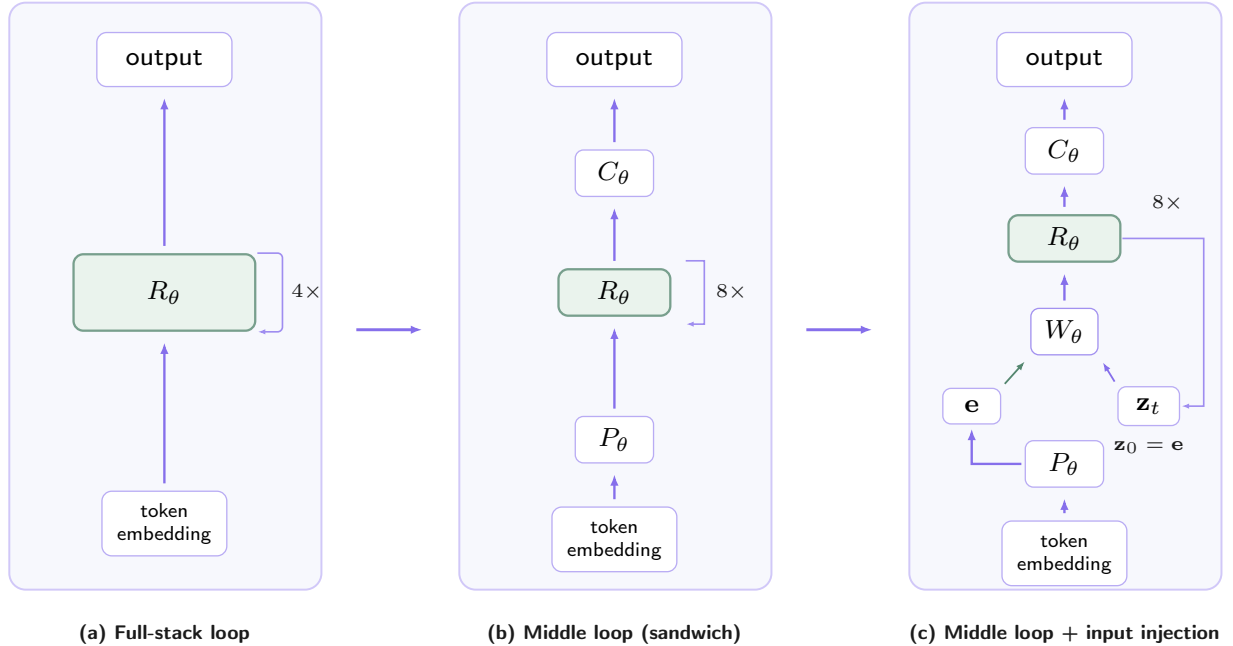

\subsubsection{Q1: Does Moving Recurrence into a Sandwich Envelope Help?}
\label{sec:q1-envelope}
\begin{questionresult}{Results}
The sandwich envelope improves multi-step answer derivation from the provided
instance, raising MATH500 by $12.00$ points and DROP by $2.61$ points. Its
gains on MATH500 and BBH-CoT persist across all four training budgets.
Knowledge-heavy tasks and tasks with explicit output or interface requirements
show no consistent gain, and some scores decline.
\end{questionresult}

Q1 isolates loop placement at fixed logical compute. The full-stack
\textit{Ouro} control reuses all 28 blocks for four passes, $R_{28}^{4}$,
whereas the sandwich control places a tied 12-block recurrent core between
eight untied prelude blocks and eight untied coda blocks,
$P_{8}R_{12}^{8}C_{8}$. We compare the two architectures at 58, 115, 230, and
460 tokens per parameter (TPP), corresponding to 42B, 84B, 168B, and 336B
training tokens for the 730M-parameter models.

\begin{table}[!htbp]
  \centering
  \footnotesize
  \setlength{\tabcolsep}{4.5pt}
  \renewcommand{\arraystretch}{1.12}
  \caption{Q1 envelope comparison at 460 TPP (730M parameters;
  336B training tokens). Feedforward columns are references; bold compares
  only the loop variants, and $\Delta$ is sandwich minus full-stack. Green
  and red text marks positive and negative changes, respectively.}
  \label{tab:q-topology-only}
  \begin{tabular*}{0.90\linewidth}{@{\extracolsep{\fill}}llcc@{\hspace{0.8em}}ccc@{}}
    \toprule
    \multicolumn{2}{@{}l}{}
      & \multicolumn{2}{c}{\textit{Feedforward reference}}
      & \multicolumn{3}{c@{}}{\textit{Loop topology}} \\
    \cmidrule(lr){3-4}\cmidrule(l){5-7}
    Group & Benchmark
      & $D_{28}$
      & $D_{112}$
      & $R_{28}^{4}$
      & $P_8R_{12}^{8}C_8$
      & $\Delta$ \\
    \midrule
    \multirow{6}{*}{\textit{General}}
      & ARC-C & 36.39 & 43.78 & \textbf{39.40} & 38.45 & \textcolor{deltalosstext}{$-0.94$} \\
      & HSwag & 41.73 & 50.28 & 44.71 & \textbf{45.01} & \textcolor{deltagaintext}{$+0.30$} \\
      & MMLU & 39.86 & 54.85 & 46.32 & \textbf{47.27} & \textcolor{deltagaintext}{$+0.95$} \\
      & BBH-CoT & 34.42 & 55.91 & 45.83 & \textbf{48.30} & \textcolor{deltagaintext}{$+2.47$} \\
      & TQA & 25.21 & 38.83 & 28.51 & \textbf{28.60} & \textcolor{deltagaintext}{$+0.09$} \\
      & DROP & 31.04 & 40.88 & 36.38 & \textbf{38.99} & \textcolor{deltagaintext}{$+2.61$} \\
    \midrule
    \multirow{2}{*}{\textit{Math}}
      & MATH500 & 12.80 & 47.60 & 20.80 & \textbf{32.80} & \textcolor{deltagaintext}{$+12.00$} \\
      & GSM8K & 29.11 & 64.22 & 54.97 & \textbf{62.02} & \textcolor{deltagaintext}{$+7.05$} \\
    \midrule
    \multirow{2}{*}{\textit{Code}}
      & HEval$^+$ & 21.34 & 42.07 & \textbf{29.88} & 28.66 & \textcolor{deltalosstext}{$-1.22$} \\
      & MBPP$^+$ & 35.19 & 51.06 & 41.01 & \textbf{43.12} & \textcolor{deltagaintext}{$+2.12$} \\
    \bottomrule
  \end{tabular*}
\end{table}

\begin{figure}[!htbp]
  \centering
  \includegraphics[width=\linewidth]{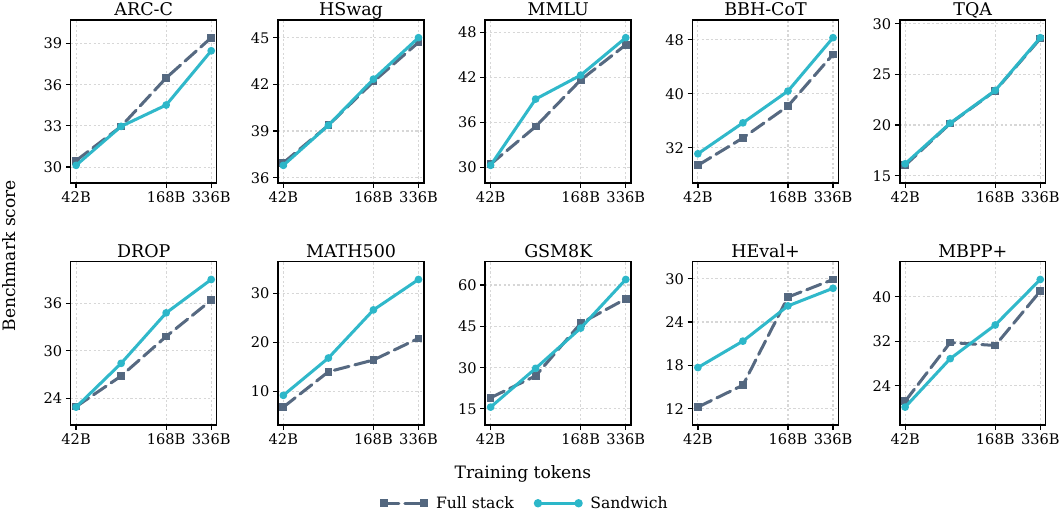}
  \caption{Evaluation trajectories for the full-stack and middle-loop sandwich variants.}
  \label{fig:q1-topology-profiles}
\end{figure}

\begin{findinganalysis}{Finding 1: The sandwich envelope improves instance-conditioned, multi-step answer derivation}
\findingstatement{Its largest and most consistent benefits occur on tasks that
require quantitative, symbolic, or heterogeneous reasoning over the supplied
input.}
\end{findinganalysis}
\analysissummary{MATH500, BBH-CoT, and DROP favor the envelope across the
training budgets, whereas the gain on GSM8K is also visible. In this
architecture, recurrence is localized to the middle of the network: only
$R_{12}$ is reapplied, and the coda decodes its hidden state only after the
final iteration. This separation may allow the recurrent core to devote more
of its capacity to iterative state refinement, while the untied prelude and
coda specialize in input encoding and output decoding. This interpretation is
consistent with prior theoretical and empirical evidence that repeated
middle-layer computation can support multi-step
reasoning~\citep{saunshi2025latentthoughts,geiping2025recurrentdepth,cai2026t2mlr}.}

\begin{findinganalysis}{Finding 2: The sandwich envelope does not consistently improve knowledge-intensive or requirement-constrained tasks}
\findingstatement{Tasks that depend heavily on stored knowledge or strict output
and interface requirements show mixed or negative changes.}
\end{findinganalysis}

\analysissummary{Knowledge-intensive benchmarks depend substantially on
information stored in the model parameters rather than evidence contained in
the evaluated instance. Code-generation tasks also require derived computation,
but success additionally depends on satisfying a precise executable
specification and producing syntactically and semantically valid outputs. The
observed pattern therefore bounds the benefit of the sandwich envelope: the
results are consistent with improved reasoning over instance-provided
information, but do not indicate better retrieval of learned knowledge or more
reliable compliance with strict output contracts. Although the envelope benefits
several reasoning benchmarks, that advantage alone is insufficient to improve
the full mixture of capabilities required for code generation.}

\subsubsection{Q2: Does Input Injection Help?}

\begin{questionresult}{Results}
Input injection yields its clearest gains on tasks that rely heavily on prompt-provided context or specifications. It improves BBH-CoT, DROP, MMLU, and both coding benchmarks, but can reduce performance on quantitative reasoning tasks, particularly in the middle-loop architecture. Despite this trade-off, the complete middle-loop model with prelude-state injection outperforms full-stack \textit{Ouro} with raw-token injection on eight of the ten benchmarks.
\end{questionresult}

Persistent input injection gives every recurrent step a direct path back to the input, reducing the need for the evolving state to carry all prompt-relevant information across iterations and potentially improving gradient propagation. Prior work on recurrent-depth and fixed-point models uses related mechanisms to keep iterative states explicitly conditioned on the input~\citep{bai2019deep,geiping2025recurrentdepth,prairie2026parcae,labovich2026stability}. We evaluate matched write-versus-no-write controls within two recurrent families, thereby isolating the effect of injection within each architecture.

Let $\mathbf x_0=E_\theta(\mathbf x)$ be the raw token-embedding stream and $\mathbf e=P_\theta(\mathbf x_0)$ the contextualized representation produced by untied prelude blocks. For a generic write source $\mathbf v$, both controls use the same diagonal operator~\citep{prairie2026parcae}
\[
D(\mathbf z_t,\mathbf v)=
\boldsymbol\alpha\odot \mathbf z_t
+ \boldsymbol\delta\odot \mathbf W_{\mathrm{in}}\mathbf v,
\qquad
\boldsymbol\delta=\operatorname{softplus}(\mathbf b_\delta),
\qquad
\boldsymbol\alpha=\exp\{-\boldsymbol\delta\odot\exp(\mathbf a)\}.
\]

The middle loop writes the fixed contextualized state $\mathbf v=\mathbf e$ before each recurrent-core pass. Full-stack \textit{Ouro} has no untied prelude, so its corresponding intervention writes the raw token embeddings $\mathbf v=\mathbf x_0$ before each full-stack pass. Each write-versus-no-write pair isolates injection within one family.

Repeated writes make every update directly dependent on $\mathbf v$, so the write source remains available after every preceding recurrent transformation. This direct path should be particularly useful when later computation must recover task instructions, contextual evidence, demonstrations, or specifications from the prompt.

\begin{table}[!htbp]
  \centering
  \footnotesize
  \setlength{\tabcolsep}{3.0pt}
  \renewcommand{\arraystretch}{1.12}
  \caption{Recurrent input-injection controls at 460 TPP (730M parameters; 336B training tokens). For the middle loop,
  $\Delta$ denotes prelude-injection minus no injection; for full-stack
  \textit{Ouro}, it denotes raw-token injection minus no injection. Boldface compares
  models within the same family, while green and red values indicate positive and
  negative changes, respectively.}
  \label{tab:q2-recurrent-input-injection}
  \begin{tabular*}{0.94\linewidth}{@{\extracolsep{\fill}}llccc@{\hspace{0.7em}}ccc@{}}
    \toprule
    \multicolumn{2}{@{}l}{}
      & \multicolumn{3}{c}{\textit{Middle loop}}
      & \multicolumn{3}{c@{}}{\textit{Full-stack \textsc{Ouro}}} \\
    \cmidrule(lr){3-5}\cmidrule(l){6-8}
    Group & Benchmark
      & no write & prelude write & $\Delta$
      & no write & raw-token write & $\Delta$ \\
    \midrule
    General & ARC-C & \textbf{38.45} & 37.60 & \textcolor{deltalosstext}{$-0.86$} & \textbf{39.40} & 39.06 & \textcolor{deltalosstext}{$-0.34$} \\
      & HSwag & 45.01 & \textbf{45.37} & \textcolor{deltagaintext}{$+0.36$} & 44.71 & \textbf{45.65} & \textcolor{deltagaintext}{$+0.94$} \\
      & MMLU & 47.27 & \textbf{49.79} & \textcolor{deltagaintext}{$+2.53$} & 46.32 & \textbf{48.11} & \textcolor{deltagaintext}{$+1.79$} \\
      & BBH-CoT & 48.30 & \textbf{54.94} & \textcolor{deltagaintext}{$+6.63$} & 45.83 & \textbf{47.63} & \textcolor{deltagaintext}{$+1.80$} \\
      & TQA & 28.60 & \textbf{29.42} & \textcolor{deltagaintext}{$+0.82$} & 28.51 & \textbf{28.53} & \textcolor{deltagaintext}{$+0.02$} \\
      & DROP & 38.99 & \textbf{40.38} & \textcolor{deltagaintext}{$+1.39$} & 36.38 & \textbf{37.29} & \textcolor{deltagaintext}{$+0.91$} \\
    \midrule
    Math & MATH500 & \textbf{32.80} & 29.20 & \textcolor{deltalosstext}{$-3.60$} & \textbf{20.80} & 19.20 & \textcolor{deltalosstext}{$-1.60$} \\
      & GSM8K & \textbf{62.02} & 59.51 & \textcolor{deltalosstext}{$-2.51$} & 54.97 & \textbf{55.50} & \textcolor{deltagaintext}{$+0.53$} \\
    \midrule
    Code & HEval$^+$ & 28.66 & \textbf{34.15} & \textcolor{deltagaintext}{$+5.49$} & 29.88 & \textbf{32.32} & \textcolor{deltagaintext}{$+2.44$} \\
      & MBPP$^+$ & 43.12 & \textbf{47.35} & \textcolor{deltagaintext}{$+4.23$} & 41.01 & \textbf{45.50} & \textcolor{deltagaintext}{$+4.50$} \\
    \bottomrule
  \end{tabular*}
\end{table}

\begin{figure}[!htbp]
  \centering
  \includegraphics[width=\linewidth]{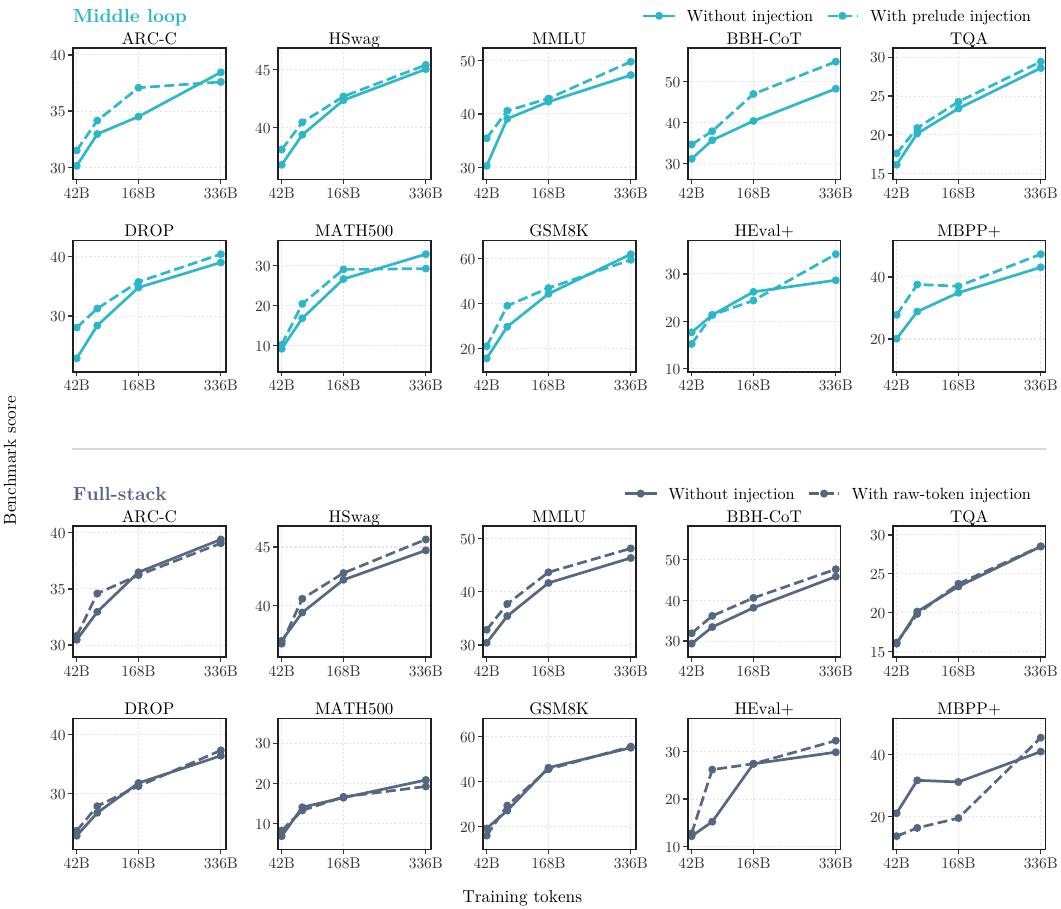}
  \caption{Evaluation trajectories over matched token budgets for input-injection ablations. The upper two rows compare the middle loop without and with prelude injection; the lower two rows compare full-stack (\textit{Ouro}-style) recurrence without and with token embedding injection.}
  \label{fig:q2-input-injection-profiles}
\end{figure}

\begin{findinganalysis}{Finding 1: Input injection primarily benefits context-dependent and specification-constrained tasks}
\end{findinganalysis}

Within the middle-loop architecture, prelude-state injection improves MMLU by 2.53 points, BBH-CoT by 6.63 points, DROP by 1.39 points, HumanEval+ by 5.49 points, and MBPP+ by 4.23 points. Raw-token injection produces gains in the same broad capability groups for full-stack \textit{Ouro}, including improvements of 1.79 points on MMLU, 1.80 points on BBH-CoT, 0.91 points on DROP, 2.44 points on HumanEval+, and 4.50 points on MBPP+.

\analysissummary{These tasks depend heavily on information provided in the evaluated instance. HumanEval$^+$ and MBPP$^+$ require generated programs to satisfy prompt-supplied specifications~\citep{chen2021evaluating,austin2021program,liu2023evalplus}; BBH-CoT provides worked examples~\citep{suzgun2023challenging}; and DROP provides an answer-bearing passage~\citep{dua2019drop}. The consistent gains across both recurrent families therefore support the interpretation that persistent injection helps later recurrent computation retain access to task-relevant contextual information.}

\begin{findinganalysis}{Finding 2: Prelude-state injection can impair quantitative reasoning}
\end{findinganalysis}

The effect of injection is not uniformly positive. In the middle-loop model, prelude-state injection reduces MATH500 by 3.60 points and GSM8K by 2.51 points. The corresponding effects are less negative in full-stack \textit{Ouro}: raw-token injection reduces MATH500 by 1.60 points and slightly improves GSM8K by 0.53 points.

This pattern suggests that repeatedly reintroducing a fixed contextualized representation may interfere with the iterative state transformations required for quantitative reasoning. One possible interpretation is that mathematical tasks benefit from allowing the recurrent state to progressively depart from the input representation as intermediate computations accumulate. Persistent prelude-state injection may instead continually pull the state back toward its
input-conditioned representation. The present experiments establish the behavioral trade-off but do not isolate its mechanism, which may depend on the write strength, injection location, or interaction between the fixed prelude state and the evolving recurrent state.

\begin{findinganalysis}{Finding 3: Despite this trade-off, the complete injected middle-loop design outperforms injected full-stack \textit{Ouro} on eight of ten benchmarks}
\end{findinganalysis}

With their respective input-conditioning mechanisms enabled, the prelude–loop–coda model outperforms full-stack \textit{Ouro} on eight of ten benchmarks, trailing only on ARC-Challenge and HellaSwag. The middle-loop model also remains stronger on MATH500 and GSM8K, even though prelude-state injection lowers its performance relative to the no-write control.

These results highlight the overall strength of the middle-loop design. The sandwich envelope and contextualized input access appear complementary on context-dependent and code-generation tasks, while the envelope advantage compensates for the negative effect of injection on quantitative reasoning. Overall, the prelude–loop–coda architecture provides the stronger foundation, although its input-injection mechanism could be further refined to better preserve mathematical reasoning performance.

\subsubsection{Q3: Do Separate Recurrent Latent States Help?}

\begin{questionresult}{Results}
Additional latent-state structure provides no consistent benefit. Random initialization produces two gains and four losses exceeding one point. The H/L recurrent-state variant produces three gains and three losses exceeding one point, with the largest degradation occurring on MATH500.
\end{questionresult}

After fixing the middle-loop topology and prelude-conditioned write, we vary
only the organization of the recurrent latent state.

Random initialization completes the controlled transformation from the
\textit{Ouro}-style design to the standard \textit{Huginn}-style
configuration. The shared H/L-state variant then tests an additional
hierarchical organization inspired by HRM-Text~\citep{wang2026hrmtext} and
TRM~\citep{jolicoeurmartineau2025morerecursivereasoningtiny}. Unlike
HRM-Text, which uses separate modules for the H and L updates, our matched
control shares a single recurrent body across both states.

We use $H=2$ high-level cycles and $L=3$ low-level updates per cycle. The
resulting schedule applies the shared recurrent body eight times---six
low-level updates and two high-level updates---matching the recurrent compute
of the single-state control. Let $\mathbf e=P_\theta(\mathbf x_0)$ be the
contextualized representation produced by the prelude; we compare three
variants:
\begin{itemize}[leftmargin=1.35em,itemsep=1pt,topsep=2pt,parsep=0pt]
  \item \textbf{Input-initialized single state:} $\mathbf z_0=\mathbf e$.
  \item \textbf{Random-initialized single state:} $\mathbf z_0=\boldsymbol{\xi}$,
    where each token state follows $\boldsymbol{\xi}\sim
    \mathcal{N}(\mathbf 0,\mathbf I_d/d)$.
  \item \textbf{Shared high/low states:} a high-level state is initialized from
    $\mathbf e$, a low-level workspace is initialized from noise, and both
    states are updated by the same recurrent body according to an H/L
    schedule.
\end{itemize}

Let
\[
  \bar R_\theta(\mathbf s;\mathbf e)
  =R_\theta(W_\theta(\mathbf s,\mathbf e))
\]
denote one application of the shared recurrent body preceded by the
prelude-conditioned input write.
\[
\begin{aligned}
  \mathbf z_{\mathrm H}^{0} &= \mathbf e,
  \qquad
  \mathbf z_{\mathrm L}^{0,0}=\boldsymbol{\xi},\quad
  \boldsymbol{\xi}\sim
  \mathcal{N}(\mathbf 0,\mathbf I_d/d), \\
  \mathbf z_{\mathrm L}^{i,j+1}
    &= \bar R_\theta(
      \mathbf z_{\mathrm L}^{i,j}+\mathbf z_{\mathrm H}^{i};
      \mathbf e),
  \qquad j=0,\ldots,L-1, \\
  \mathbf z_{\mathrm H}^{i+1}
    &= \bar R_\theta(
      \mathbf z_{\mathrm H}^{i}+\mathbf z_{\mathrm L}^{i,L};
      \mathbf e),
  \qquad
  \mathbf z_{\mathrm L}^{i+1,0}=\mathbf z_{\mathrm L}^{i,L},
  \qquad i=0,\ldots,H-1 .
\end{aligned}
\]
Thus, each low-level update receives the current high-level state through
additive conditioning before entering the shared recurrent body, while each
high-level update analogously receives the final low-level state from the
corresponding cycle.

\usetikzlibrary{fit,shapes.geometric}
\begin{figure}[!htbp]
  \centering
  \resizebox{0.76\linewidth}{!}{%
  \begin{tikzpicture}[font=\sffamily\scriptsize, >=Latex]
    \tikzset{
      hl panel/.style={
        draw=ifmpurpletext!30,
        fill=ifmblueaccent!26,
        rounded corners=6pt,
        line width=0.65pt
      },
      hl subpanel/.style={
        draw=ifmpurpletext!16,
        fill=white,
        rounded corners=5pt,
        line width=0.45pt
      },
      hl state/.style={
        draw=ifmpurpletext!45,
        fill=white,
        rounded corners=3pt,
        minimum width=1.05cm,
        minimum height=0.42cm,
        align=center,
        font=\scriptsize
      },
      hl state wide/.style={
        hl state,
        minimum width=1.35cm,
        minimum height=0.44cm
      },
      hl call/.style={
        draw=loopfindingtext!65,
        fill=loopfindingback,
        rounded corners=4pt,
        minimum width=1.54cm,
        minimum height=0.52cm,
        align=center,
        line width=0.74pt,
        font=\scriptsize
      },
      hl add/.style={
        draw=ifmpurpletext!55,
        fill=white,
        circle,
        minimum size=0.42cm,
        inner sep=0pt,
        align=center,
        font=\scriptsize
      },
      hl source/.style={
        draw=ifmpurpletext!45,
        fill=white,
        rounded corners=2pt,
        minimum width=0.42cm,
        minimum height=0.34cm,
        inner sep=0.8pt,
        align=center,
        font=\scriptsize
      },
      hl switch/.style={
        draw=ifmpurpletext!55,
        fill=white,
        diamond,
        aspect=1.45,
        minimum width=0.62cm,
        minimum height=0.46cm,
        inner sep=0pt,
        align=center,
        font=\scriptsize
      },
      hl flow/.style={
        -{Latex[length=1.16mm,width=0.84mm]},
        line width=0.64pt,
        draw=ifmpurpletext!78,
        shorten >=0.7pt,
        shorten <=0.5pt,
        line cap=round
      },
      hl loop/.style={
        -{Latex[length=0.95mm,width=0.68mm]},
        line width=0.58pt,
        draw=ifmpurpletext!62,
        shorten >=0.5pt,
        shorten <=0.35pt,
        line cap=round
      },
      hl init/.style={
        -{Latex[length=0.92mm,width=0.66mm]},
        line width=0.48pt,
        draw=ifmpurpletext!58,
        shorten >=0.45pt,
        shorten <=0.35pt,
        line cap=round
      },
      hl note/.style={
        font=\sffamily\tiny,
        text=ifmtext,
        align=center
      }
    }

    \path[use as bounding box] (0,0) rectangle (9.15,5.35);
    \node[
      hl panel,
      minimum width=8.85cm,
      minimum height=4.78cm,
      anchor=south west
    ] at (0.15,0.38) {};
    \node[
      hl subpanel,
      minimum width=6.50cm,
      minimum height=1.70cm,
      anchor=south west
    ] at (1.24,1.15) {};

    \node[hl state] (zh) at (1.98,3.52)
      {$\mathbf z_{\mathrm H}^{i}$};
    \node[hl state wide] (zl) at (1.98,2.02)
      {$\mathbf z_{\mathrm L}^{i,j}$};
    \node[hl add] (merge) at (3.05,2.85) {$+$};
    \node[hl call] (rbar) at (4.70,2.85)
      {$\bar R_\theta(\mathord\cdot;\mathbf e)$};
    \node[hl switch] (sw) at (6.16,2.85) {$\sigma$};
    \node[hl state] (zhnext) at (7.72,3.62)
      {$\mathbf z_{\mathrm H}^{i+1}$};
    \node[hl state wide] (zlnext) at (7.72,1.90)
      {$\mathbf z_{\mathrm L}^{i,j+1}$};

    \draw[hl flow] (zh.east) -- (merge.north |- zh.east) -- (merge.north);
    \draw[hl flow] (zl.east) -- (merge.south |- zl.east) -- (merge.south);
    \draw[hl flow] (merge.east) -- (rbar.west);
    \draw[hl flow] (rbar.east) -- (sw.west);
    \draw[hl flow] (sw.north east) -- (6.68,3.62) -- (zhnext.west);
    \draw[hl flow] (sw.south east) -- (6.68,1.90) -- (zlnext.west);
    \node[hl note, fill=ifmpurpletext!4, inner sep=0.8pt]
      at (6.58,3.30) {$\sigma=\mathrm H$};
    \node[hl note, fill=white, inner sep=0.8pt]
      at (6.58,2.40) {$\sigma=\mathrm L$};

    \draw[hl loop] (zhnext.north) -- (7.72,4.20) -- (1.98,4.20) -- (zh.north);
    \node[hl note, fill=ifmblueaccent!26, inner sep=1.0pt]
      at (4.58,4.38) {each cycle: $L$ lows $\rightarrow$ one high};

    \node[hl source] (prelude) at (0.70,3.52) {$\mathbf e$};
    \node[hl note] at (0.70,3.18)
      {$\mathbf z_{\mathrm H}^{0}=\mathbf e$};
    \draw[hl init] (prelude.east) -- (zh.west);

    \node[hl source] (noise) at (0.70,2.02) {$\boldsymbol{\xi}$};
    \node[hl note] at (0.70,1.68)
      {$\boldsymbol{\xi}\sim\mathcal{N}$};
    \draw[hl init] (noise.east) -- (zl.west);
    \draw[hl loop] (zlnext.south) -- (7.72,1.34) -- (1.98,1.34) -- (zl.south);
    \node[hl note, fill=white, inner sep=0.8pt]
      at (4.58,1.14) {$L$ low updates};

    \node[font=\sffamily\bfseries\scriptsize, text=ifmtext]
      at (4.58,0.12) {Low/high latent states};
  \end{tikzpicture}%
  }
  \caption{Shared high/low latent-state update. The same recurrent body
  alternates low-state refinements with a high-state update.}
  \label{fig:q3-low-high-update}
\end{figure}
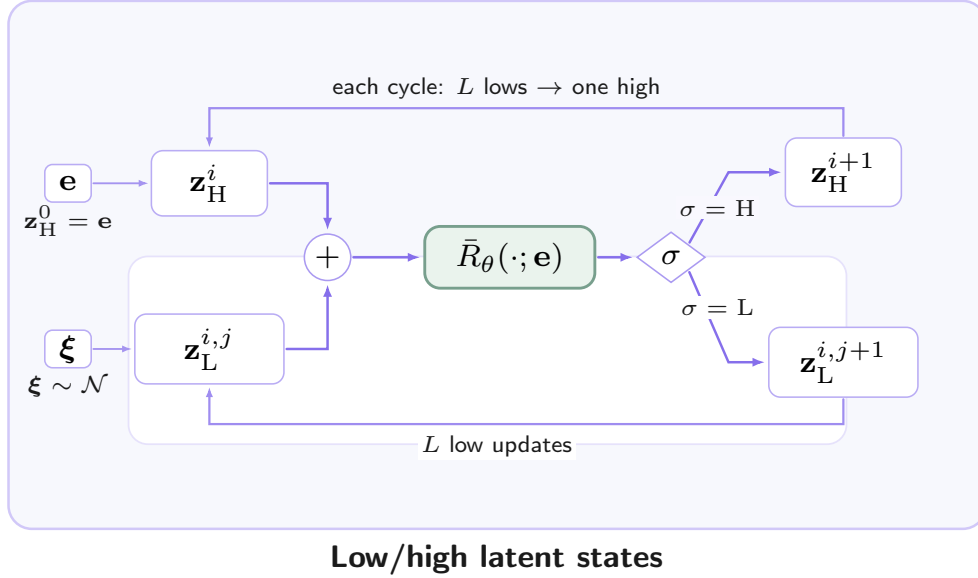

\begin{table}[!htbp]
  \centering
  \footnotesize
  \setlength{\tabcolsep}{4.5pt}
  \renewcommand{\arraystretch}{1.12}
  \caption{Matched recurrent-state comparison at 460 TPP (730M parameters; 336B training tokens). Bold marks the best available score.
  $\Delta$ is shared high/low (\(\mathrm{H/L}\)) minus the input-initialized
  single state (\(\mathbf z_0=\mathbf e\)). Green and red text mark positive and
  negative changes.}
  \label{tab:q3-latent-state}
  \begin{tabular*}{0.82\linewidth}{@{\extracolsep{\fill}}llccc@{\hspace{0.8em}}c@{}}
    \toprule
    \multicolumn{2}{@{}l}{}
      & \multicolumn{3}{c}{\textit{Latent state}}
      & \multicolumn{1}{c@{}}
        {\textit{\(\mathrm{H/L}-(\mathbf z_0=\mathbf e)\)}} \\
    \cmidrule(lr){3-5}
    Group & Benchmark
      & $\mathbf z_0=\mathbf e$
      & $\mathbf z_0\sim\mathcal{N}$
      & $\mathrm{H/L}$
      & $\Delta$ \\
    \midrule
    \multirow{6}{*}{\textit{General}}
      & ARC-C & 37.60 & \textbf{40.94} & 39.83 & \textcolor{deltagaintext}{$+2.23$} \\
      & HSwag & 45.37 & 45.47 & \textbf{45.90} & \textcolor{deltagaintext}{$+0.53$} \\
      & MMLU & \textbf{49.79} & 48.50 & 49.49 & \textcolor{deltalosstext}{$-0.30$} \\
      & BBH-CoT & 54.94 & 54.43 & \textbf{55.12} & \textcolor{deltagaintext}{$+0.18$} \\
      & TQA & 29.42 & 28.97 & \textbf{30.17} & \textcolor{deltagaintext}{$+0.75$} \\
      & DROP & 40.38 & 40.63 & \textbf{42.18} & \textcolor{deltagaintext}{$+1.80$} \\
    \midrule
    \multirow{2}{*}{\textit{Math}}
      & MATH500 & \textbf{29.20} & 27.40 & 22.00 & \textcolor{deltalosstext}{$-7.20$} \\
      & GSM8K & 59.51 & \textbf{60.73} & 57.32 & \textcolor{deltalosstext}{$-2.19$} \\
    \midrule
    \multirow{2}{*}{\textit{Code}}
      & HEval$^+$ & 34.15 & 32.32 & \textbf{35.37} & \textcolor{deltagaintext}{$+1.22$} \\
      & MBPP$^+$ & \textbf{47.35} & 45.24 & 44.71 & \textcolor{deltalosstext}{$-2.64$} \\
    \bottomrule
  \end{tabular*}
\end{table}

\begin{figure}[!htbp]
  \centering
  \includegraphics[width=\linewidth]{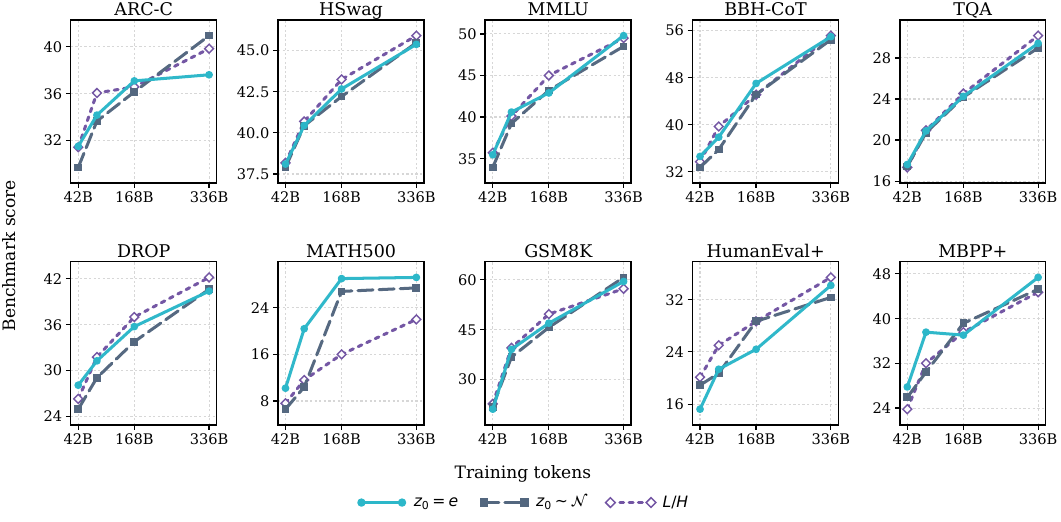}
  \caption{Evaluation benchmark trajectories for the three latent-state
  organizations.}
  \label{fig:q3-latent-state-profiles}
\end{figure}

\begin{findinganalysis}{Finding 1: Random initialization is not a necessary ingredient for looped language models}
\end{findinganalysis}

Random initial states are widely used in iterative models~\citep{wang2025hrm,huang2026equilibrium,geiping2025recurrentdepth,baek2026generative,feinashley2026solveloop}, partly because they are thought to encourage path independence on algorithmic equilibrium tasks~\citep{anil2022pathindependent}. We argue, however, that random initialization should instead be viewed as a task- and objective-dependent inductive bias, rather than as a universally beneficial design choice.

Relative to direct initialization with $\mathbf z_0=\mathbf e$, random initialization improves ARC-Challenge by 3.34 points and GSM8K by 1.22 points. However, it reduces performance by more than one point on MMLU, MATH500, HumanEval$^{+}$ and MBPP$^{+}$, while the differences on the remaining four benchmarks are within one point.

To our knowledge, prior work has not isolated the effect of random initialization in a matched, large-scale, finitely unrolled language model with persistent input injection. Across our evaluations, random initialization produces improvements exceeding one point on two benchmarks but degradations exceeding one point on four, providing no evidence of a consistent cross-task advantage in this setting. Direct initialization performs better on six of the ten benchmarks and also avoids the additional computation associated with sampling a separate initial state. These results should not, however, be interpreted as evidence against random initialization in general: our evaluations do not directly measure multi-start path independence or extrapolation to recurrent depths beyond those used during training.

\begin{findinganalysis}{Finding 2: A shared-module H/L hierarchy provides no consistent benefit}
\end{findinganalysis}

HRM~\citep{wang2025hrm} introduces an H/L hierarchy through a brain-inspired
separation of timescales. A fast L module repeatedly refines a local state
under a fixed H-level context, after which a slower H module updates that
context and initiates a new phase of low-level refinement. This alternating
schedule is intended to sustain effective computational depth. HRM-Text adapts
the same two-cycle, three-L-update schedule to language modeling and combines
it with MagicNorm, warmup credit assignment, and PrefixLM training~\citep{wang2026hrmtext}.

Existing evidence for the value of the H/L hierarchy is mixed. An L-only model
matches HRM on Sudoku~\citep{ge2025criticalhrm}, whereas TRM reports
improvements from combining two recurrent states with shared weights on
Sudoku-Extreme~\citep{jolicoeurmartineau2025morerecursivereasoningtiny}. At
language-model scale, HRM-Text compares a separate-module HRM architecture
with a shared-weight TRM-style alternative~\citep{wang2026hrmtext}. However,
these comparisons also vary the update rule, effective depth, parameter count,
or computational budget, making it difficult to isolate the contribution of
the state hierarchy itself.

Our control holds the recurrent body, number of recurrent-body applications,
input injection, parameter count, and training budget fixed. It introduces
only a second recurrent state and the H/L update schedule, with the same
recurrent module shared across both states. Under this matched setting, the
hierarchy yields gains exceeding one point on three benchmarks and losses
exceeding one point on three, while substantially degrading MATH500. Thus, the
additional state and scheduling complexity provides no consistent aggregate
benefit when the H and L updates share a recurrent module.

This conclusion is specific to the shared-module setting. It does not rule out
benefits from the separately parameterized H and L modules used by HRM-Text. A
closer HRM-Text-style variant with distinct modules for the two update streams
is still under evaluation and is not included in the present comparison.

\subsection{Summary}

\begin{figure}[!htbp]
  \centering
  \includegraphics[width=\linewidth]{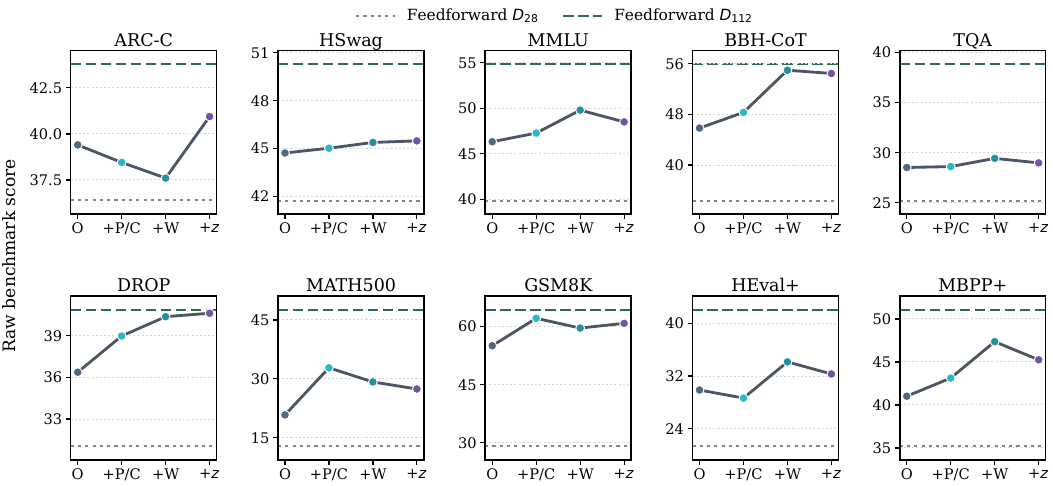}
  \caption{Raw benchmark scores along the cumulative dense construction path
  at 336B training tokens. Starting from full-stack \textit{Ouro}
  (\(\mathrm{O}\)), the stages add the untied prelude/coda sandwich envelope
  (\(+\mathrm{P/C}\)), prelude-conditioned input injection
  (\(+\mathrm{W}\)), and a separate
  random-initialized recurrent state (\(+z\)), yielding complete
  \textit{Huginn}-style looped models. Horizontal lines show the matched \(D_{28}\) and
  \(D_{112}\) feedforward references.}
  \label{fig:ouro-huginn-path-336b}
\end{figure}

The sandwich envelope and persistent input injection account for the broadest gains along the controlled construction path, whereas introducing a randomly or separately initialized recurrent state provides no consistent additional benefit. Figure~\ref{fig:ouro-huginn-path-336b} shows that the
sandwich envelope and persistent input injection drive the gains, while a
separate random-initialized latent state provides no consistent further
benefit.

\FloatBarrier

The stepwise results distinguish the contributions of the three design changes. The sandwich envelope produces the clearest gains on mathematical and multi-step reasoning tasks. Persistent input injection produces its largest additional gains on BBH-CoT and code generation, while also improving DROP and MMLU. 

Introducing a separately random-initialized recurrent state has mixed effects and reverses some of the preceding gains on several benchmarks. Under this construction order, the results suggest that the broad advantage of \textit{Huginn} over \textit{Ouro} is driven mainly by localized recurrence and repeated access to a contextualized input representation.

\section{Transfer to MoE Models}
\label{sec:moe-transfer}

\begin{questionresult}{Results}
\textbf{1)} The \textit{Huginn}-style MoE retains its dense-setting advantage over
\textit{Ouro}. \textbf{2)} It achieves more balanced expert utilization, and
iteration-specific expert selection contributes to performance. \textbf{3)}
Although the feedforward MoE leads overall, \textit{Huginn} surpasses it on DROP
and GSM8K, matches it on MATH500, and substantially narrows the
recurrent-to-feedforward gap relative to the dense setting.
\end{questionresult}

\subsection{Configurations for MoE Comparison}

We replace the FFN in every physical layer of the two complete loop
architectures with a mixture-of-experts (MoE) layer. The \textit{Ouro}-style
model retains full-stack recurrence and an input-initialized recurrent state.
The \textit{Huginn}-style model retains the sandwich envelope,
prelude-conditioned input writes, and random state initialization.

Both recurrent models contain 28 routed physical layers with 25 experts, top-2
routing, and expert width 2432. Each has 8.0B resident parameters and 793.9M
active parameters per physical pass. Unrolling the recurrent computation yields
approximately 3.2B active parameter applications per token. A 112-layer
feedforward model without cross-layer weight sharing, denoted
$D_{112}^{\mathrm{MoE}}$, provides a reference matched in logical depth and
active compute, with 32.0B resident parameters. All MoE models use dot-product load-balancing loss without router bias.

\subsection{\textit{Huginn}-Style MoE versus \textit{Ouro}}

\begin{table}[!htbp]
  \centering
  \footnotesize
  \setlength{\tabcolsep}{4.5pt}
  \renewcommand{\arraystretch}{1.12}
  \caption{Benchmark performance of looped MoE models and the feedforward MoE reference after training on 500B tokens. The looped models contain 8.0B resident and 793.9M active parameters per physical pass, corresponding to approximately 630 training tokens per active parameter. Bold compares only the recurrent rows; $\Delta$ is \textit{Huginn MoE} minus \textit{Ouro MoE}; green and red mark positive and negative changes, respectively.}
  \label{tab:allmoe-final-budget}
  \begin{tabular*}{0.86\linewidth}{@{\extracolsep{\fill}}llc@{\hspace{0.65em}\vrule\hspace{0.65em}}cc@{\hspace{0.8em}}c@{}}
    \toprule
    \multicolumn{2}{@{}l}{}
      & \multicolumn{3}{c}{\textit{MoE architecture}}
      & \multicolumn{1}{c@{}}{} \\
    \cmidrule(lr){3-5}
    Group & Benchmark
      & Feedforward $D_{112}^{\mathrm{MoE}}$
      & \textsc{Ouro} $R_{28}^{4}$
      & \textsc{Huginn} $P_8R_{12}^{8}C_8$
      & $\Delta$ \\
    \midrule
    \multirow{6}{*}{\textit{General}}
      & ARC-C & 53.22 & \textbf{51.33} & 50.47 & \textcolor{deltalosstext}{$-0.86$} \\
      & HSwag & 57.30 & 54.82 & \textbf{55.10} & \textcolor{deltagaintext}{$+0.28$} \\
      & MMLU & 67.13 & 64.03 & \textbf{66.81} & \textcolor{deltagaintext}{$+2.78$} \\
      & BBH-CoT & 71.76 & \textbf{69.05} & 68.85 & \textcolor{deltalosstext}{$-0.20$} \\
      & TQA & 59.02 & 50.98 & \textbf{51.04} & \textcolor{deltagaintext}{$+0.07$} \\
      & DROP & 56.81 & 57.70 & \textbf{59.65} & \textcolor{deltagaintext}{$+1.95$} \\
    \midrule
    \multirow{2}{*}{\textit{Math}}
      & MATH500 & 53.40 & 49.80 & \textbf{53.40} & \textcolor{deltagaintext}{$+3.60$} \\
      & GSM8K & 80.82 & 78.92 & \textbf{83.62} & \textcolor{deltagaintext}{$+4.70$} \\
    \midrule
    \multirow{2}{*}{\textit{Code}}
      & HEval$^+$ & 56.10 & 48.17 & \textbf{49.39} & \textcolor{deltagaintext}{$+1.22$} \\
      & MBPP$^+$ & 60.85 & 58.99 & \textbf{59.52} & \textcolor{deltagaintext}{$+0.53$} \\
    \bottomrule
  \end{tabular*}
\end{table}

\begin{findinganalysis}{Finding 1: The \textit{Huginn}-style advantage over \textit{Ouro} transfers to MoE models}
\findingstatement{\textit{Huginn} scores higher on eight of ten benchmarks; five
of these gains exceed one point, with its largest gains on GSM8K ($+4.70$) and
MATH500 ($+3.60$).}
\end{findinganalysis}

Table~\ref{tab:allmoe-final-budget} shows that the endpoint difference is
not uniform across tasks. Within the general benchmarks, the clearest gains for
\textit{Huginn} are on MMLU ($+2.78$) and DROP ($+1.95$). Its leads on
HellaSwag and TriviaQA are only $0.28$ and $0.07$ points, respectively, while
\textit{Ouro} is higher on ARC-Challenge by $0.86$ points and on BBH-CoT by
$0.20$.

The separation is larger on mathematical reasoning. \textit{Huginn} improves
over \textit{Ouro} by $3.60$ points on MATH500 and $4.70$ on GSM8K. The code
differences are smaller: $+1.22$ on HumanEval$^+$ and $+0.53$ on MBPP$^+$.
Overall, the transfer advantage is concentrated on mathematical reasoning, with
additional gains on MMLU and DROP, while performance on the remaining
benchmarks is broadly comparable.

\subsection{Routing Diagnostics}

We compare normalized load-balancing loss across the two recurrent models and
the feedforward reference. Uniform routing has a loss of one; lower values
therefore indicate more even expert utilization.

\begin{figure}[!htbp]
  \centering
  \includegraphics[width=0.82\linewidth]{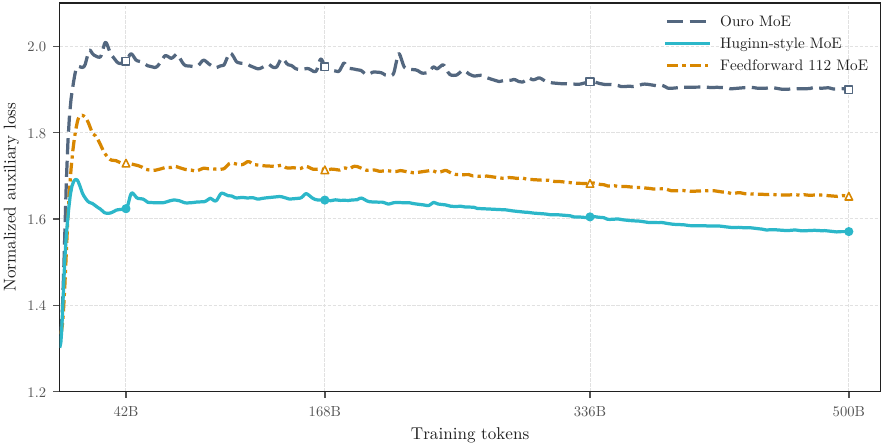}
  \caption{Normalized MoE load-balancing loss across training for the recurrent
  models and the 112-layer feedforward MoE reference (lower is better). Curves
  show 0.5B-token bin medians after robust smoothing. At 500B tokens,
  \textsc{Huginn} remains below both comparators, and the feedforward reference
  lies between \textsc{Huginn} and \textsc{Ouro}.}
  \label{fig:moe-aux-loss}
\end{figure}

\FloatBarrier

\begin{findinganalysis}{Finding 2}
\findingstatement{The \textit{Huginn} MoE exhibits the lowest load-balancing loss
at 500B tokens: $1.571$, compared with $1.652$ for the feedforward reference and
$1.899$ for \textit{Ouro}. This indicates more even expert utilization.}
\end{findinganalysis}

\begin{figure}[!htbp]
  \centering
  \includegraphics[width=0.78\linewidth]{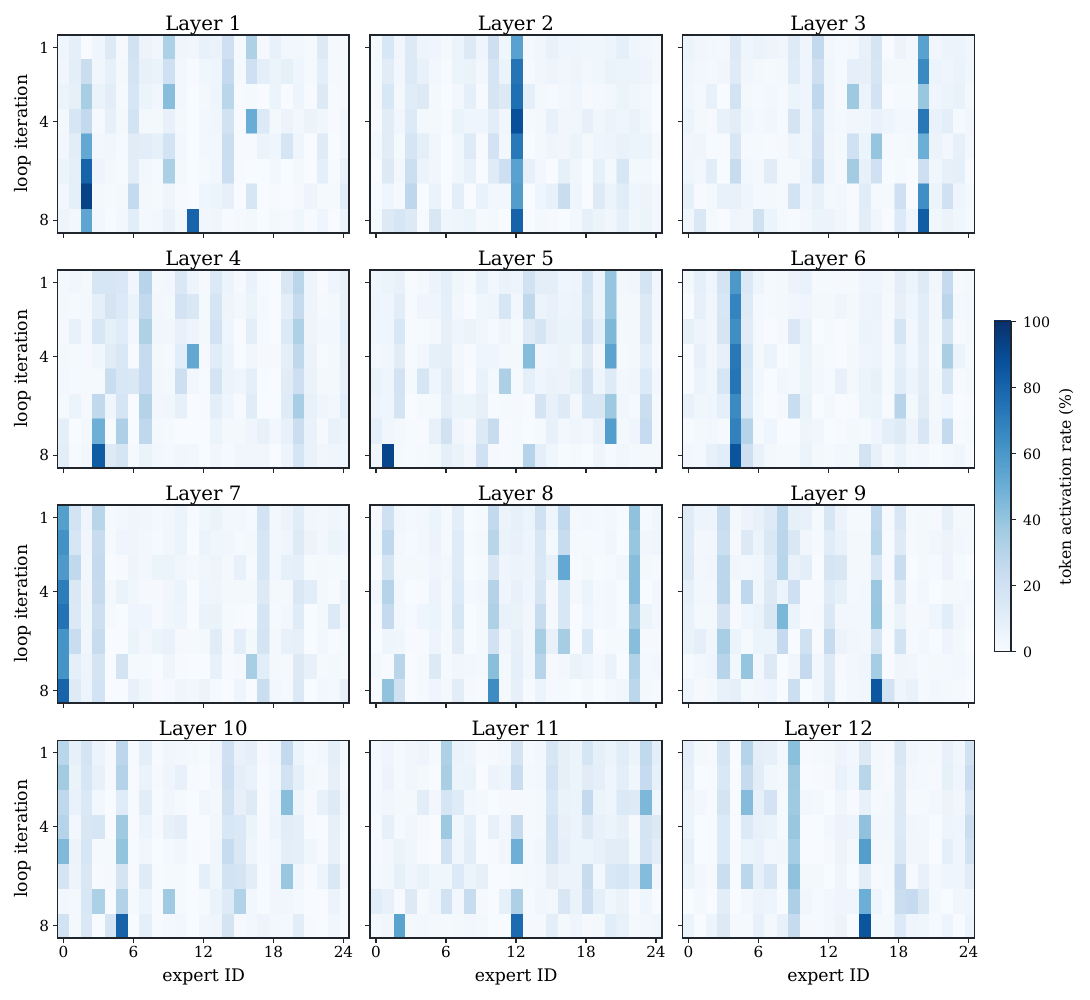}
  \caption{Iteration-dependent top-2 expert activation at the 500B-token
  checkpoint across all 12 recurrent layers, computed on a 48-input probe set
  with eight examples each from
  ARC-Challenge, MMLU, GSM8K, MATH500, HumanEval, and MBPP. Each panel shows the
  fraction of valid tokens activating each expert over the eight loop
  iterations, using one shared color scale.}
  \label{fig:moe-iteration-routing}
\end{figure}

\begin{figure}[!htbp]
  \centering
  \includegraphics[width=0.86\linewidth]{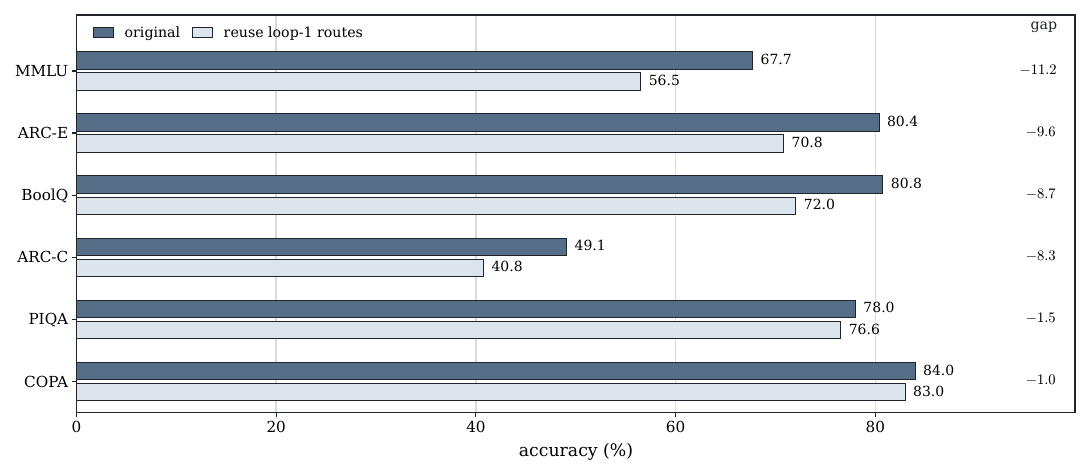}
  \caption{Original accuracy and accuracy after iterations 2--8 reuse the
  loop-1 top-2 expert identities on MMLU, ARC-Easy, BoolQ, ARC-Challenge, PIQA,
  and COPA. The gap at right is original minus frozen-route accuracy.}
  \label{fig:moe-frozen-route-accuracy}
\end{figure}

\analysissummary{We find that the \textit{Huginn} MoE exhibits a substantially
lower load-balancing loss, indicating more even expert utilization.
Figure~\ref{fig:moe-iteration-routing} further shows that the
distribution of top-2 expert activations varies across loop iterations. This
variation is functionally important: forcing iterations 2--8 to reuse the
expert identities selected at iteration 1, while retaining iteration-specific
mixture weights, reduces accuracy on all six evaluated tasks
(Figure~\ref{fig:moe-frozen-route-accuracy}). This intervention provides
causal evidence that iteration-specific expert selection contributes to model
performance. Appendix~\ref{app:moe-routing-probe} describes the probe set and
the intervention.}

\subsection{\textit{Huginn}-MoE versus Feedforward MoE}

We compare \textit{Huginn} with the 112-layer feedforward MoE under the same
logical depth and active-compute budget. We further compare the
loop-to-feedforward performance gaps in the dense and MoE settings.

\begin{figure}[!htbp]
  \centering
  \includegraphics[width=\linewidth]{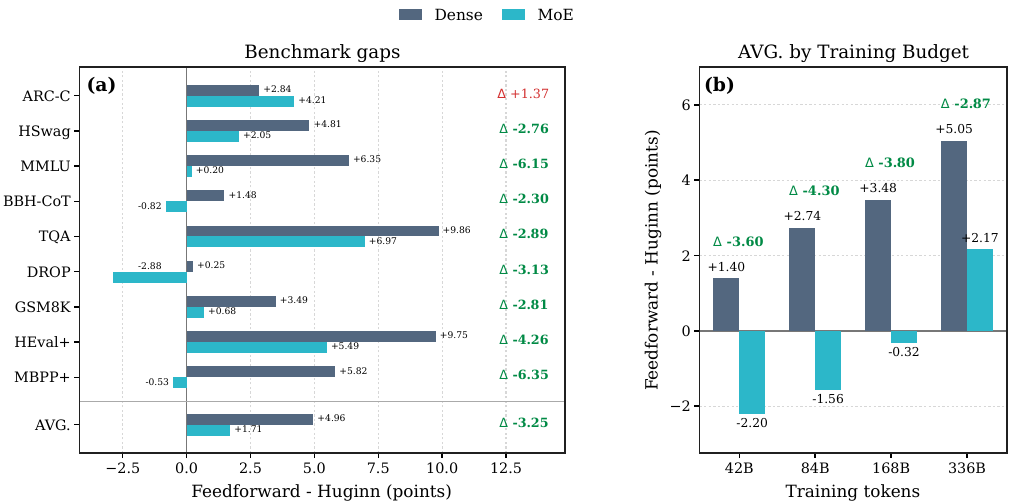}
  \caption{Dense and MoE gaps to the 112-layer feedforward references.
  For each model family $f\in\{\mathrm{dense},\mathrm{MoE}\}$,
  $g_f=S(D_{112}^{f})-S(\textsc{Huginn}^{f})$; positive values favor the
  feedforward reference.  Panel~(a) compares the individual benchmark gaps at
  the shared endpoint.  Each green or red $\Delta$ is
  $g_{\mathrm{MoE}}-g_{\mathrm{dense}}$, so negative values indicate that MoE
  narrows the recurrent-to-feedforward gap.  Panel~(b) follows the same gap
  comparison over training on the fixed seven-task intersection (ARC-C,
  HSwag, MMLU, BBH-CoT, TQA, DROP, and HumanEval$^+$).}
  \label{fig:dense-moe-unrolled-gap}
\end{figure}

\begin{findinganalysis}{Finding 3: MoE substantially narrows the performance gap between \textit{Huginn} and its feedforward counterpart}
\findingstatement{The feedforward MoE leads on seven of ten benchmarks, whereas
\textit{Huginn} leads on DROP and GSM8K and matches it on MATH500. Across
MATH500, GSM8K, HumanEval$^+$, and MBPP$^+$, \textit{Huginn} averages $61.48$,
compared with $62.79$ for the feedforward reference.}
\end{findinganalysis}

\analysissummary{Figure~\ref{fig:dense-moe-unrolled-gap} compares the dense and
MoE performance gaps between \textit{Huginn} and their respective 112-layer
feedforward references. In panel~(a), the MoE gap is smaller on eight of the
nine directly comparable benchmarks, reducing the mean gap from $4.96$ to $1.71$
points. Panel~(b) shows that this pattern holds throughout training: when
evaluated on the same seven benchmarks, the MoE gap is smaller at all four
checkpoints.}

\subsection{Summary}

\begin{itemize}
  \item First, the \textit{Huginn}-style MoE outperforms \textit{Ouro} on eight of
  ten benchmarks, with its largest gains on GSM8K and MATH500 and additional
  gains exceeding one point on MMLU, DROP, and HumanEval$^+$. This shows that
  the advantage observed in the dense setting transfers to MoE models.
  \item Second, \textit{Huginn} achieves more balanced expert utilization, and
  intervention experiments show that iteration-specific expert selection
  contributes meaningfully to performance.
  \item Third, MoE substantially narrows the performance gap between
  \textit{Huginn} and its feedforward counterpart. Although the feedforward
  baseline leads overall, \textit{Huginn} outperforms it on DROP and GSM8K,
  matches it on MATH500, and reduces the mean gap from 4.96 points in the dense
  setting to 1.71 points in the MoE setting.
\end{itemize}

\FloatBarrier

\section{Future Work}
\label{sec:future-work}

Two extensions are needed to establish how broadly these results hold. First,
a small-to-large scaling ladder should repeat the controlled comparisons
across resident parameter counts and training budgets. The current experiments
characterize finite-budget behavior at one practical model scale. Establishing
a scaling law requires matched comparisons across model sizes and training
budgets. Extending these comparisons to models with hundreds of billions of
parameters remains beyond our current compute budget. Second, multi-seed
replications should quantify the seed-to-seed uncertainty of the reported
differences.

The MoE study also leaves a component-attribution question. It compares
complete \textit{Ouro}- and \textit{Huginn}-style configurations, which change
the iteration envelope, input injection, and state initialization together.
Matched MoE ablations should vary these axes separately. Further experiments
could combine the shared H/L state with expert routing and repeat the
route-freezing intervention across checkpoints and generation tasks.

\section{Conclusions}
\label{sec:conclusion}

We compare full-stack and middle-loop language models by disentangling three design choices that standard \textit{Ouro}- and \textit{Huginn}-style recipes vary simultaneously: the iteration envelope, input injection, and latent-state organization. Under matched parameter scale, logical depth, and training-token budget, we find that the prelude--loop--coda sandwich architecture improves performance on mathematical reasoning, context extraction, and other reasoning-intensive tasks. Input injection primarily benefits knowledge retrieval and context access, while contributing little to quantitative reasoning. Random state initialization yields mixed results, whereas the shared H/L hierarchy provides no consistent benefit.

Our MoE transfer study evaluates the two complete architectures in a modern sparse-model setting and shows that the advantages of the \textit{Huginn}-style architecture persist. The \textit{Huginn} MoE achieves more balanced expert utilization and approaches---and on some tasks surpasses---the performance of its feedforward MoE counterpart, despite being trained from scratch under the same data budget and matched memory and FLOP constraints.

More broadly, the architectural design space of looped models remains underexplored and calls for substantially more controlled ablation studies. Meaningful comparisons therefore require carefully designed, apples-to-apples evaluations that match parameter scale, training budget, inference cost, and evaluation protocol. Many important questions remain open, including how loop architectures interact with conditional sparsity and how architecture--system co-design can further advance their performance--efficiency Pareto frontier.

\bibliographystyle{unsrtnat}
\bibliography{references}

\appendix
\beginappendix
\setcounter{subsection}{0}
\setcounter{subsubsection}{0}
\renewcommand{\thesubsection}{\Alph{subsection}}
\renewcommand{\thesubsubsection}{\thesubsection.\arabic{subsubsection}}
\subsection{Model and Training Configuration}
\label{app:model-training}

\textbf{Transformer block.} All compared models use the same decoder-only
PreNorm backbone with hidden size 1536, RMSNorm before each branch
($\epsilon=10^{-6}$), and separate attention and feed-forward residuals.
Attention uses 24 query heads and 6 key--value heads, head dimension 64,
causal attention, and RoPE over all head dimensions with base $10^6$; we do
not apply query/key normalization. The dense branch is a SwiGLU MLP of width
4352. Each MoE variant retains this block structure and replaces only the
dense feed-forward branch with 25 width-2432 SwiGLU experts, sigmoid top-2
routing, and no shared experts. Table~\ref{tab:model-capacity-accounting}
gives the MoE geometry and parameter accounting.

\begin{table}[!htbp]
  \centering
  \scriptsize
  \setlength{\tabcolsep}{3.6pt}
  \renewcommand{\arraystretch}{1.08}
  \caption{Model geometry and non-embedding parameter accounting for the MoE models.}
  \label{tab:model-capacity-accounting}
  \resizebox{\linewidth}{!}{%
  \begin{tabular}{@{}l c c c c c c c c c c@{}}
    \toprule
    & \multicolumn{3}{c}{Geometry}
    & \multicolumn{3}{c}{MoE}
    & \multicolumn{4}{c}{Parameters} \\
    \cmidrule(lr){2-4}\cmidrule(lr){5-7}\cmidrule(l){8-11}
    Model
      & Phys.
      & Topology
      & MoE phys.
      & $E$
      & Top-$k$
      & $d_{\mathrm{expert}}$
      & Real
      & Active
      & Loop-equiv.
      & Unrolled active \\
    \midrule
    \multicolumn{11}{@{}l}{\textit{Feedforward reference}} \\
    Feedforward MoE
      & 112
      & $D_{112}$
      & 112
      & 25 & 2 & 2432
      & 32.0B & 3.2B & n/a & n/a \\
    \midrule
    \multicolumn{11}{@{}l}{\textit{Looped models}} \\
    \textsc{Huginn} MoE
      & 28
      & $P_8R_{12}^{8}C_8$
      & 28
      & 25 & 2 & 2432
      & 8.0B & 793.9M & 32.0B & 3.2B \\
    \textsc{Ouro} MoE
      & 28
      & $R_{28}^{4}$
      & 28
      & 25 & 2 & 2432
      & 8.0B & 793.9M & 32.0B & 3.2B \\
    \bottomrule
  \end{tabular}%
  }
\end{table}

\textbf{Optimization.} We train on TxT360 with sequence length 8192 and
AdamW. We sweep learning rates in
$\{4\times10^{-4},6\times10^{-4},8\times10^{-4}\}$ and report the best
setting, use 200 warmup steps followed by cosine decay, and set weight decay
to 0.1. Every run uses an effective global batch of 512 sequences, or
$512\times8192=4{,}194{,}304$ tokens per optimizer step.

\subsection{Evaluation Protocol}
\label{app:evaluation-protocol}

Every reported score follows the task-specific prompt, cap, and metric listed
below. Likelihood tasks score all answer choices without generation.
Generation tasks use greedy single-sample decoding, no chat template, and
fixed demonstrations. Evaluation batch size and cache use vary only for
throughput.

\begin{table}[!htbp]
  \centering
  \footnotesize
  \setlength{\tabcolsep}{3.5pt}
  \renewcommand{\arraystretch}{1.08}
  \caption{Task-level evaluation settings for the reported benchmark results.}
  \label{tab:evaluation-protocol}
  \begin{tabularx}{\linewidth}{@{}lYccr@{}}
    \toprule
    Task & Split (examples) & Shots & Metric & Input / output cap \\
    \midrule
    ARC-C
      & Shared local test (1,165)
      & 0
      & $\mathrm{Acc}_{\mathrm{tok}}$
      & 512 / -- \\
    HellaSwag
      & Validation (10,042)
      & 0
      & Acc
      & 256 / -- \\
    MMLU
      & Test, 57 subjects (14,042)
      & 5
      & Micro Acc
      & 2,048 / -- \\
    BBH-CoT
      & Full test (6,511)
      & 3
      & EM
      & 4,096 / 1,024 \\
    TriviaQA
      & Local full test (11,313)
      & 5
      & F1
      & 512 / 24 \\
    DROP
      & Test (9,535)
      & 3
      & F1
      & 2,500 / 64 \\
    GSM8K
      & Test (1,319)
      & 8
      & Answer Acc
      & 2,048 / 512 \\
    MATH500
      & HuggingFaceH4 MATH-500 (500)
      & 4
      & EM
      & 4,096 / 2,048 \\
    HumanEval$^+$
      & EvalPlus HumanEval (164)
      & 0
      & pass@1
      & 1,024 / 512 \\
    MBPP$^+$
      & Canonical EvalPlus (378)
      & 3
      & pass@1
      & 3,096 / 1,024 \\
    \bottomrule
  \end{tabularx}
\end{table}

The two cap values are the maximum input and generated-token lengths.
``--'' denotes a likelihood task with no decoded output. Each task family uses
the following prompt and stopping rule:
\begin{itemize}[leftmargin=1.5em,itemsep=0.12em,topsep=0.3em,parsep=0pt]
  \item \textbf{Choice likelihood.} ARC-C and HellaSwag score all four answer
    choices. MMLU uses fixed dev demonstrations.
  \item \textbf{Chain-of-thought generation.} BBH-CoT uses fixed
    category-matched demonstrations and stops at the next \texttt{Q:} or blank
    block. GSM8K uses fixed demonstrations and stops at the next \texttt{Q:}
    or \texttt{A:}.
  \item \textbf{Short-answer generation.} TriviaQA uses fixed short-answer
    demonstrations, and DROP uses fixed passage-QA demonstrations. Both stop
    at the next question, answer, or newline.
  \item \textbf{Mathematics.} MATH500 uses the official HuggingFaceH4
    MATH-500 set and fixed Minerva demonstrations, stopping at EOS or at the
    next \texttt{Problem:}.
  \item \textbf{Code.} HumanEval$^+$ completes the supplied function and stops
    at EOS. MBPP$^+$ uses fixed code demonstrations and the
    \texttt{[DONE]} stop.
\end{itemize}
Prompts longer than the input cap are left-truncated. In MMLU, the
2,048-token cap can remove the earliest demonstrations for long questions,
while the current question and all answer choices remain intact.

\paragraph{Reporting rules.}
BBH-CoT uses the three-shot chain-of-thought protocol; answer-only values are
not mixed into the displayed comparisons. ARC-C reports
answer-token-normalized choice accuracy, whereas HellaSwag reports raw
highest-likelihood choice accuracy. A small set of retained TriviaQA traces
predates newline stopping; it uses the same five-shot prompt, 24-token cap,
and F1 scorer. HumanEval$^+$ and MBPP$^+$ are scored with the EvalPlus extra
tests. The HumanEval$^+$ values use 512-token generation traces; a paired
1,024-token check at the matched 168B checkpoints produced the same pass@1
values. Most MATH500 rows use the standard Minerva prompt with EOS stopping;
three Q3 rows use a pinned Minerva prompt variant with a decoded
next-\texttt{Problem:} stop. For the routing intervention in
Appendix~\ref{app:moe-routing-probe}, MMLU is the unweighted mean of the 57
subject accuracies; the benchmark tables use example-level micro accuracy.

\subsection{MoE Routing Probe}
\label{app:moe-routing-probe}

The routing diagnostic uses the 500B-token \textit{Huginn}-style MoE
checkpoint with 12 recurrent physical layers, eight iterations, 25 experts per
layer, and top-2 routing. The 48-input probe set contains eight examples each
from ARC-Challenge, MMLU, GSM8K, MATH500, HumanEval, and MBPP; inputs are
truncated to 1,024 tokens. The switch statistic compares the top-1 identity
for the same token and physical layer at adjacent iterations. It excludes
padding and the final input position, leaving 47,928 token--layer decisions
per transition. Across adjacent iterations, the mean top-1 expert switch rate
is $49.4\%$, and every transition has a nonzero switch rate.

The heatmap in Figure~\ref{fig:moe-iteration-routing} aggregates these
valid positions by physical layer, iteration, and expert. Each panel fixes one
recurrent layer; columns index experts, rows index iterations, and cell
intensity is the fraction of valid token decisions that select an expert in
either top-2 slot.

For the frozen-identity intervention, iteration 1 runs normally. At iterations
2--8, each token and physical layer reuses iteration 1's top-2 expert indices.
The model recomputes and renormalizes the current hidden state's scores for
those fixed identities. The intervention therefore removes identity switching
while retaining iteration-specific mixture weights. We compare normal and
frozen routing on complete, otherwise matched multiple-choice evaluations of
ARC-Easy, ARC-Challenge, BoolQ, COPA, PIQA, and all 57 MMLU subjects. Freezing
route identities lowers accuracy on all six tasks
(Figure~\ref{fig:moe-frozen-route-accuracy}). The drop is largest on
MMLU ($-11.2$ points), ARC-Easy ($-9.6$), BoolQ ($-8.7$), and ARC-Challenge
($-8.3$); PIQA and COPA decrease by $1.5$ and $1.0$ points.

\subsection{Related Work on Recurrent Depth and Sparse Routing}
\label{app:related-work}

We organize prior work by its role in our comparisons. Complete recurrent
architectures motivate the attribution problem, mechanism studies motivate the
controlled axes, and sparse models define the closest compute-matched
comparators.

\subsubsection{Complete Recipes Leave Architectural Attribution Unresolved}

\textit{Huginn} shows that a recurrent middle block can scale latent
test-time computation without generating additional reasoning
tokens~\citep{geiping2025recurrentdepth}. \textit{Ouro} develops full-stack
looping as a pretraining and scaling direction and combines it with learned
depth allocation~\citep{zhu2025ouro}. Together, they establish middle-loop and
full-stack recurrence as viable pretraining recipes. Work on
iso-depth scaling, capability generalization, and stable looped scaling
studies how these models behave across depth and scale
~\citep{schwethelm2026recurrence,kohli2026loop,prairie2026parcae}.

Endpoint comparisons between complete recipes cannot determine whether a
difference comes from the prelude--loop--coda envelope, the per-iteration
input path, the initial recurrent state, or another training choice. Our
controlled dense path addresses this attribution by holding resident scale,
logical block executions, data, checkpoint
tokens, and evaluation protocol fixed while changing the iteration envelope,
input write, and latent-state organization in separate comparisons. The design
conclusions therefore apply to this matched regime.

\subsubsection{Input and State Design Shape Recurrent Dynamics}

\paragraph{Input injection changes the recurrent update.}
Deep-equilibrium models use input injection to make an implicit state depend
on the input~\citep{bai2019deep}. In looped language models, Parcae traces
training instability to the injection parameterization and constrains its
spectral behavior~\citep{prairie2026parcae}. A fixed-point analysis separates
reachability, input dependence, and geometry, and studies recall placement on
structured tasks~\citep{labovich2026stability}. STARS regularizes recurrent
dynamics for test-time depth, while residual-scaling work derives a
loop-specific scaling rule for correlated updates across repeated
applications~\citep{yang2026stars,wang2026residual}. Across these settings,
loop count must be interpreted together with the update rule.

\paragraph{State channels can induce distinct recurrent roles.}
Mechanistic analysis connects recurrent block size, input injection, and
normalization to cyclic fixed-point dynamics~\citep{blayney2026mechanistic}.
On structured reasoning tasks, learned memory tokens can provide a necessary
working-state channel, and state-identity or asymmetric-input signals can
induce distinct roles in a shared recurrent
block~\citep{sapunov2026memory,shen2026tworoles}. CART supplies a negative
result: in its small-scale language-model study, prelude depth matters more
than loop count, and varying the inference loop count away from its trained
value degrades performance~\citep{capps2026cart}.

These studies motivate the axes in Section~\ref{sec:preliminaries} across
fixed-point theory, structured reasoning, mechanistic probes, and language
models at different scales. Our matched pretraining controls study
architectural attribution in one regime. The input-write ablations compare
write versus no write within each family; raw-token and contextualized writes
remain coupled to different recurrence envelopes.

\subsubsection{Sparse Routing Adds Conditional Capacity to Recurrent Depth}

Reusing a dense block increases computation without increasing its stored
capacity. Sparse Universal Transformer and MoEUT combine shared depth with
conditional computation. Their evaluations target Universal Transformers,
leaving the recurrent-family comparison open~\citep{tan2023sparseut,csordas2024moeut}.
More recent work reports that
looped MoE models benefit from routing divergence across passes and can expose
useful early exits~\citep{lee2026sparse}. Mixture of Universal Experts shares
an expert pool across layers to convert physical depth into virtual
width~\citep{chen2026moue}. It studies repeated expert reuse; our study
compares the \textit{Huginn} and \textit{Ouro} recurrent families.

\paragraph{LoopMoE provides the closest feedforward comparison.}
LoopMoE constructs a prelude--loop--coda MoE and matches a vanilla MoE in total
parameters, per-token FLOPs, and the active attention-to-FFN ratio
\citep{chen2026loopmoe}. Its 3B model contains six unique layers: two prelude
layers, a two-layer block executed four times, and two coda layers. Both it and
the feedforward reference therefore execute twelve logical layers. The
reported 0.6B ``physical active'' count de-duplicates shared parameters; the
compute comparison uses a 0.8B active budget. At 200B training tokens,
LoopMoE leads the matched vanilla MoE on eight of nine reported benchmarks.
Its 9B comparison is a 100B-token early checkpoint. The 3B result provides
controlled evidence that recurrence can improve a matched sparse baseline.
The persistence of the 9B gap through full training remains unresolved.

\paragraph{Our MoE comparison tests recurrent-family ordering.}
We study whether the ordering between two complete recurrent architectures
persists after routed experts replace their feed-forward sublayers.
The \textit{Huginn}- and \textit{Ouro}-style models are resident- and
active-matched and trained through 500B tokens. The 112-layer feedforward
model matches the logical active-compute path and stores more parameters,
providing a compute-matched reference. Our route-freezing intervention tests
whether the \textit{Huginn}-style model uses iteration-dependent expert
choices. The multi-scale recurrent--feedforward comparison and component-level
MoE ablations remain open.

\end{document}